\documentclass[journal,onecolumn]{IEEEtran}
\usepackage[total={6in,9in}]{geometry}

\usepackage{amsfonts,amssymb,amsthm,latexsym}
\usepackage[hidelinks]{hyperref}
\usepackage[cmex10]{amsmath}
\usepackage{graphicx,psfrag,xspace}
\hypersetup{colorlinks = false, citecolor = blue}
\usepackage[usenames,dvipsnames,table]{xcolor}
\usepackage{float, subfig}
\usepackage{setspace} 

\usepackage{mathtools} 
\usepackage{import} 

\renewcommand{\hat}{\widehat}
\renewcommand{\tilde}{\widetilde}
\renewcommand{\bar}{\overline}

\newcommand{\whp}{w.h.p.\@\xspace}
\newcommand{\kmeans}{$k$-means\xspace} 

\newcommand{\scp}[2]{\langle #1, #2 \rangle}
\newcommand{\iid}{%
  \ifmmode
  \mathrm{i.i.d.}%
  \else%
  i.i.d.\@\xspace%
  \fi%
}
\newcommand{\wpone}{%
  \ifmmode
  \mathrm{w.p.1}%
  \else%
  w.p.1\@\xspace%
  \fi%
}

\DeclareMathOperator*{\argmin}{arg\,min}
\DeclareMathOperator*{\argmax}{arg\,max}

\DeclareMathOperator*{\vect}{vec}
\renewcommand{\Vec}[1]{\bm{#1}} 

\DeclareMathOperator*{\sign}{sign}

\DeclareMathOperator{\rank}{rank}
\DeclareMathOperator{\MMD}{MMD}
\newcommand{\smid}{|\,}

\newcommand{\cA}{\mathcal{A}}

\newcommand{\cH}{\mathcal{H}}

\newcommand{\cN}{\mathcal{N}}

\newcommand{\cX}{\mathcal{X}}

\newcommand{\bb}{\mathbb}

\newcommand{\bE}{\mathbb{E}}

\newcommand{\bR}{\mathbb{R}}

\DeclarePairedDelimiter{\prt}{(}{)}

\DeclarePairedDelimiter{\norm}{\lVert}{\rVert}

\usepackage{tabularx} 
\usepackage{booktabs} 
\usepackage{cleveref}
\usepackage{amsthm}
\usepackage[most]{tcolorbox}
\usepackage{caption}
\tcbset{texthighlight/.style={%
    enhanced,
    breakable,
    parbox=false,
    boxrule=0mm,
    borderline={0.2mm}{0mm}{darkgray!85!white,dashed},
    arc=0.2mm, outer arc=0mm,
    boxsep=1mm,
    left=1mm,
    right=1mm,
    top=0pt,
    bottom=1pt,
    before skip=10pt, 
    after skip=15pt 
      }%
    }%

    \newtcolorbox[auto counter]{highlight}[2][]{texthighlight, nameref={#2},
   colback=highlightbg,colframe=darkgray!15!darkgray,fonttitle=\bfseries,title={[Box\,\thetcbcounter] #2},#1}

\newcommand{\ybref}[1]{Box\,\ref{#1}} 
\theoremstyle{plain}
\newtheorem*{remark*}{Remark}

\usepackage{ifthen} 
\newif\ifarxiv
\newif\ifrecompiletikz

\arxivtrue 

\usepackage{tikz,pgfplots}
\usetikzlibrary{arrows,arrows.meta,positioning,decorations.pathmorphing,decorations.pathreplacing,calc,patterns,fit}
\usetikzlibrary{calc,arrows,positioning,matrices,datavisualization,datavisualization.formats.functions}

\usepackage{enumitem} 
\setitemize{noitemsep,topsep=2pt,parsep=2pt,partopsep=0pt}
\setenumerate{noitemsep,topsep=2pt,parsep=2pt,partopsep=0pt}

\makeatletter
\newcommand\subparagraph{%
  \@startsection{subparagraph}{5}
  {\parindent}
  {3.25ex \@plus 1ex \@minus .2ex}
  {-1em}
  {\normalfont\normalsize\bfseries}}
\makeatother
\usepackage{titlesec}
\titleformat{\section}{\normalfont\fontsize{12}{15}\scshape\bfseries}{\thesection}{1em}{}
\titleformat{\subsection}{\normalfont\fontsize{10}{12}\scshape\bfseries}{\thesubsection}{1em}{}

\ifarxiv

\def\thesection{\arabic{section}}                
\def\thesubsection{\thesection.\arabic{subsection}}

\else

\def\thesection{\arabic{section}}                
\def\thesubsection{\thesection.\arabic{subsection}}

\fi

\usepackage[style=ieee,url=false,isbn=false,doi=false,style=numeric-comp,giveninits,backend=biber,sorting=none,maxnames=10]{biblatex}

\AtEveryBibitem{\clearfield{pages}} 
\ifarxiv
    \newcommand{\ocite}[2][]{\cite[#1]{#2}} 
    \newcommand{\modifiedtext}{\color{black}} 
    \newcommand{\new}[1]{\textcolor{black}{#1}} 
    \newcommand{\finalrevisions}[1]{#1} 
\else
    \newcommand{\ocite}[2][]{\unskip\unpenalty} 
    \newcommand{\modifiedtext}{\color{black}} 
    \newcommand{\new}[1]{\textcolor{black}{#1}} 
    \newcommand{\finalrevisions}[1]{\textcolor{blue}{#1}} 
\fi

\newif\ifcomment
\commenttrue

\ifcomment
\usepackage[normalem]{ulem} 
\newcommand{\unmodifiedtext}{\color{black}}
\newcommand{\del}[1]{\textcolor{red}{\sout{#1}}}

\newcommand{\AC}[1]{\textcolor{MidnightBlue}{\textbf{\scriptsize [Antoine: #1]}}}
\newcommand{\mAC}[1]{\marginpar{\tiny{\AC{#1}}}}
\newcommand{\RG}[1]{\textcolor{Orange}{\textbf{\scriptsize [R\'emi: #1]}}}
\newcommand{\mRG}[1]{\marginpar{\tiny{\RG{#1}}}}
\newcommand{\NK}[1]{\textcolor{Plum}{\textbf{\scriptsize [Nicolas: #1]}}}
\newcommand{\mNK}[1]{\marginpar{\tiny{\NK{#1}}}}
\newcommand{\LJ}[1]{\textcolor{TealBlue}{{\bfseries \scriptsize [Laurent: #1]}}}
\newcommand{\mLJ}[1]{\marginpar{\tiny{\LJ{#1}}}}
\newcommand{\PS}[1]{\textcolor{OliveGreen}{\textbf{\scriptsize [Phil: #1]}}}
\newcommand{\mPS}[1]{\marginpar{\tiny{\PS{#1}}}}
\newcommand{\VS}[1]{\textcolor{PineGreen}{\textbf{\scriptsize [Vincent: #1]}}}
\newcommand{\mVS}[1]{\marginpar{\tiny{\VS{#1}}}}
\else
\newcommand{\modifiedtext}{}
\newcommand{\unmodifiedtext}{}
\newcommand{\new}[1]{#1}
\newcommand{\del}[1]{}
\newcommand{\sout}[1]{}

\newcommand{\AC}[1]{}
\newcommand{\mAC}[1]{}
\newcommand{\RG}[1]{}
\newcommand{\mRG}[1]{}
\newcommand{\NK}[1]{}
\newcommand{\mNK}[1]{}
\newcommand{\LJ}[1]{}
\newcommand{\mLJ}[1]{}
\newcommand{\PS}[1]{}
\newcommand{\mPS}[1]{}
\newcommand{\VS}[1]{}
\newcommand{\mVS}[1]{}
\fi

\definecolor{orangeFig2}{HTML}{EE8C54}
\definecolor{blueFig2}{HTML}{5594E1} 

\colorlet{highlightbg}{yellow!5!white} 
\colorlet{colDs}{blueFig2}
\colorlet{colDsLL}{colDs!15!white}
\colorlet{colSk}{Green}
\colorlet{colSkLL}{colSk!10!white}
\colorlet{colLn}{orangeFig2} 
\colorlet{colLnLL}{colLn!15!white}
\colorlet{colAdv}{Red}

\definecolor{coD}{HTML}{404040}
\colorlet{coDl}{coD!78} 

\newcommand{\ds}{\cX}                   

\newcommand{\vc}{\Vec{c}}

\newcommand{\vt}{\Vec{t}}

\newcommand{\vw}{\Vec{w}}
\newcommand{\vx}{\Vec{x}}
\newcommand{\vy}{\Vec{y}}
\newcommand{\ts}{\Vec{z}}       
\newcommand{\empts}{\tilde{\Vec{z}}} 
\newcommand{\psk}{\tilde{\Vec{s}}}     
\newcommand{\skop}{\cA}

\newcommand{\thetabf}{{\boldsymbol{\theta}}}
\newcommand{\Thetabf}{{\boldsymbol{\Theta}}}

\newcommand{\noise}{\Vec{e}}

\usepackage{bm}
\newcommand{\fmap}{\bm{\Phi}}
\newcommand{\cexp}{\mathrm{RF}} 

\newcommand{\cl}{\mathcal}
\newcommand{\bs}{\boldsymbol}
\newcommand{\ie}{\emph{i.e.},\xspace}
\newcommand{\eg}{\emph{e.g.},\xspace}
\newcommand{\cf}{\emph{cf.}}

\newcommand{\Exp}{\mathbb{E}}
\newcommand*\dif{\mathop{}\!\mathrm{d}}
\newcommand{\mubf}{\boldsymbol{\mu}}
\newcommand{\Sigmabf}{\boldsymbol{\Sigma}}
\newcommand{\Phibf}{\boldsymbol{\Phi}}
\newcommand{\tran}{^\top}

\newcommand{\cbf}{\Vec{c}}
\newcommand{\Abf}{\Vec{A}}

\newcommand{\Kbf}{\Vec{K}}
\newcommand{\Ibf}{\Vec{I}}
\newcommand{\pbf}{\Vec{p}}

\newcommand{\Rbf}{\Vec{R}}
\newcommand{\ubf}{\Vec{u}}

\newcommand{\Wbf}{\Vec{W}}

\newcommand{\xbf}{\Vec{x}}
\newcommand{\ybf}{\Vec{y}}
\newcommand{\wbf}{\Vec{w}}

\newcommand{\linop}{\Wbf}
\newcommand{\regmat}{\thetabf}
\renewcommand{\jmath}{\mathsf{j}} 

\begin{document}
\setlength{\arraycolsep}{0.8mm}

\ifarxiv
\title{Sketching Datasets for Large-Scale Learning (long version)}
\fi

\author{R{\'e}mi Gribonval, Antoine Chatalic, Nicolas Keriven, \protect\\
Vincent Schellekens, Laurent Jacques, and Philip Schniter%
\thanks{R. Gribonval is with Univ Lyon, Inria, CNRS, ENS de Lyon, UCB Lyon 1, LIP UMR 5668, F-69342, Lyon, France; remi.gribonval@inria.fr. A. Chatalic is with Univ Rennes, Inria, CNRS, IRISA, F-35042 Rennes Cedex, France; antoine.chatalic@irisa.fr. N. Keriven is with CNRS, GIPSA-lab, Grenoble INP, UGA, Grenoble, France; nicolas.keriven@gipsa-lab.grenoble-inp.fr. V. Schellekens and L. Jacques are with ISPGroup/ICTEAM, UCLouvain, Belgium; \{vincent.schellekens,laurent.jacques\}@uclouvain.be. P. Schniter is with The Ohio State University, Columbus, OH, USA; schniter.1@osu.edu}}

\maketitle
\thispagestyle{plain}\pagestyle{plain}

\ifarxiv
    \onehalfspacing
\fi
        
\ifarxiv
\begin{abstract}
This article considers 
\modifiedtext ``compressive learning,'' \unmodifiedtext an approach to large-scale machine learning where datasets are massively compressed before learning (\eg clustering, classification, or regression) is performed. 
In particular, a ``sketch'' is first constructed by computing carefully chosen nonlinear random features (\eg random Fourier features) and averaging them over the whole dataset. 
Parameters are then learned from the sketch, without access to the original dataset.
This article surveys the current state-of-the-art in 
\modifiedtext compressive \unmodifiedtext 
learning, 
including the main concepts and algorithms, their connections with established signal-processing methods,  existing theoretical guarantees---on both information preservation and privacy preservation, and important open problems. 
\end{abstract}
\fi



\recompiletikzfalse

Big data can be a blessing:
with very large training datasets it becomes possible to perform complex learning tasks with unprecedented accuracy. 
Yet, this improved performance comes at the price of enormous computational challenges.
Thus, one may wonder:
Is it possible to leverage the information content of huge datasets while keeping computational resources under control? 
Can this also help solve some of the privacy issues raised by large-scale learning? 
This is the ambition of 
\modifiedtext \emph{compressive learning, }\unmodifiedtext
where the dataset is massively compressed {\em before} learning.
Here, a ``sketch'' is first constructed by computing carefully chosen nonlinear random features (\eg random Fourier features) and averaging them over the whole dataset. 
Parameters are then learned from the sketch, without access to the original dataset.
This article surveys the current state-of-the-art in 
\modifiedtext compressive \unmodifiedtext learning,
including the main concepts and algorithms; their connections with established signal-processing methods; existing theoretical guarantees, on both information preservation and privacy preservation; and important open problems. 
\ifarxiv
It is an extended version of \cite{gribonval2021sketching} with additional references and more in-depth discussions on a variety of topics. 
\fi

\section{Introduction to 
\modifiedtext Compressive \unmodifiedtext
Learning} 

\begin{figure}[t]
        \newcommand{\figrecoveredparams}{\tilde{\Vec{\theta}}} 
        \begin{center}
			\ifrecompiletikz
				\input{fig-framework}
			\else
				\includegraphics[width=\textwidth]{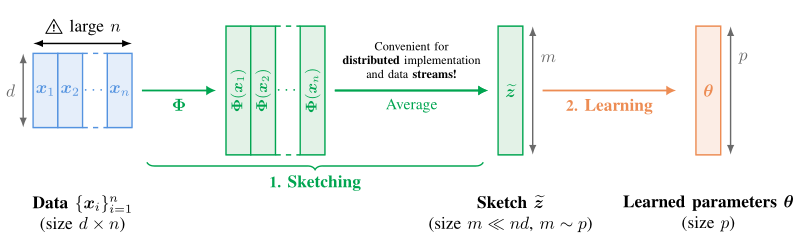}
			\fi
        \end{center}
        \caption{\small Overview of sketching and parameter learning.}
        \label{fig:framework}
\end{figure}

\begin{figure}[t]
        \begin{center}
   \ifarxiv             \includegraphics[width=\textwidth]{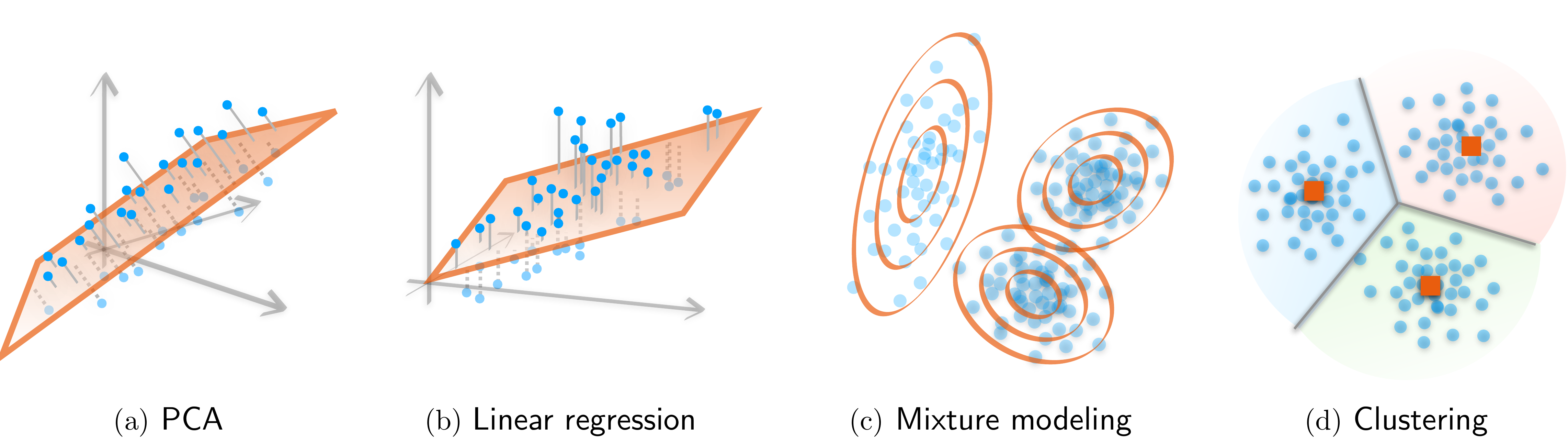}
      \fi
     \end{center}
        \caption{\small Schematic representation of the four running examples covered by this paper. In the four figures, each $\vx_i$ is associated with a blue colored disk, hence the training collection correspond to a point cloud, and the orange color geometrically represents the learned parameters. 
        \ifarxiv (a) \else \fi
        PCA learns the principal $k$-dimensional subspace of the dataset for some $k \leq d$. 
        \ifarxiv (b) \else \fi
        Linear regression fits observed data (the blue dot heights) as a linear model of the inputs (here, the 2-D horizontal coordinates). For least squares, 
        this amounts to minimizing the square of the differences between these data and the linear predictions. 
        \ifarxiv (c) \else \fi
        In Gaussian mixture modeling, we learn the set of parameters (mixture weight, mean, and covariance) characterizing each Gaussian term of the mixture, which probability level sets are displayed here as orange ellipses.
        \ifarxiv (d) \else \fi
        Clustering methods (such as \kmeans) learn a set of centroids (the orange squares) defining a Voronoi partition grouping together similar data samples. 
        }
        \label{fig:tasks}
\end{figure}

The overall principle of 
\modifiedtext compressive \unmodifiedtext
learning is summarized in Fig.~\ref{fig:framework}.
During the \emph{sketching phase}, a potentially huge collection of $d$-dimensional data vectors $\{\vx_i\}_{i=1}^n$ is summarized into a single $m$-dimensional vector $\empts$, called the ``sketch''. 
The sketch is constructed by transforming each data vector and then averaging the results:
\begin{equation}
\label{eq:sketch-def}
\empts := \frac{1}{n} \sum_{i=1}^n \fmap(\vx_i).  
\end{equation}
Next, during the \emph{learning phase}, 
an estimate $\tilde{\Vec{\theta}}$ of some
essential statistical parameters $\thetabf$ of the dataset are extracted from the sketch $\empts$.
These parameters are application dependent.
For example, as illustrated in Fig.~\ref{fig:tasks}, they could represent principal data subspaces (in subspace fitting problems),
prediction weights (in estimation and regression problems),
distributional parameters (in density estimation problems), or centroids (in clustering problems); we give detailed examples below. 
Python notebooks that illustrate examples from the paper are available at \href{https://gitlab.inria.fr/SketchedLearning/spm-notebook}{https://gitlab.inria.fr/SketchedLearning/spm-notebook}. 

The transformation $\fmap(\cdot)$, known as the ``feature map,'' is generally non-linear and randomized. 
Although the use of $\fmap(\cdot)$ is related to ``kernel methods'' in machine learning 
(\eg~\cite{perez2004kernel,scholkopf2002learning}\ocite{rosenberg2009multiview,jenssen2013entropy,arenas2013kernel}), as will be discussed later,
the act of sketching~\eqref{eq:sketch-def} also includes averaging over the $n$ data samples.
The advantages of 
\modifiedtext compressive \unmodifiedtext
learning are the following.
\begin{enumerate}
\item By choosing a sketch of dimension $m\ll nd$, the data gets massively compressed.
This has obvious advantages for storage and transfer.
\item Sketching can speed up the learning phase, whose complexity becomes independent of the cardinality, $n$, of the original dataset. 
This enables one to handle massive datasets while keeping computational resources under control.
\item Sketching can preserve privacy: the transformation $\fmap(\cdot)$ can be chosen so that individual-user information is lost while aggregate-user information is preserved.
\item The sketching mechanism in~\eqref{eq:sketch-def} is well matched to distributed implementations and streaming 
scenarios: the sketch of a concatenation of datasets is a simple mean of the sketches of those datasets.
\end{enumerate}

\section{Illustration using Four Worked Examples}  
To illustrate the 
\modifiedtext compressive\unmodifiedtext-learning framework and discuss the essential aspects of it, we now outline four canonical examples of machine learning tasks to which sketching can be readily applied. 
See Fig.~\ref{fig:tasks} for an illustration.

\paragraph{Principal component analysis (PCA)} 
PCA seeks to find the linear subspace of a fixed dimension $k<d$ that best fits the $d$-dimensional data $\{\xbf_i\}_{i=1}^n$ in the least-squares (LS) sense.
In this case, \finalrevisions{we assume the data to be centered so that} the target parameters $\thetabf$ can be described by a $k$-dimensional orthonormal basis $\{\ubf_\ell\}_{\ell=1}^k$ that maximizes $\sum_{\ell=1}^k\sum_{i=1}^n |\ubf_\ell\tran\xbf_i |^2$.

It is well known that one solution is given by the $k$ principal eigenvectors of the 
empirical autocorrelation matrix 
$\hat{\Rbf} = \frac{1}{n}\sum_{i=1}^n \xbf_i \xbf_i\tran$.
This $\hat{\Rbf}$ can be interpreted as a sketch of the form~\eqref{eq:sketch-def} 
that uses the feature map $\fmap(\xbf)=\vect(\xbf\xbf\tran)=(x_1x_1, x_2x_1,\,\cdots, x_dx_d)\tran$ of dimension $m=d^2$.
\new{Here and in the sequel, the $\vect(\cdot)$ operator vectorizes a matrix by stacking its columns.}
As soon as the cardinality $n$ of the dataset exceeds the dimension $d$, the dimension of this sketch is smaller than $nd$, the total size of the dataset. Using techniques from matrix completion and compressive matrix sensing, the sketch dimension can be further reduced \cite{FouRau13}\ocite{halko2011finding}. For example, one can sketch via~\eqref{eq:sketch-def} using 
$\Phibf(\xbf) = \big((\wbf_{1}\tran\vx)^{2},\dots,(\wbf_{m}\tran\vx)^{2}\big)\tran$, where the $d$-dimensional vectors $\wbf_{j}$, $1 \leq j \leq m$ are the rows of a tall random matrix $\Wbf$ of size $m \times d$. For $m$ on the order of $kd < d^{2}$, low-rank matrix reconstruction techniques can estimate the $k$-dimensional principal subspace of $\hat{\Rbf}$ with provable accuracy~\cite{gribonval2017}.

\paragraph{LS linear regression}
\label{sec:least-squares-ls}

Suppose that the data vectors take the form $\xbf_i=[\xbf_{1i}\tran,\xbf_{2i}\tran]\tran$, and our goal is to linearly predict the $d_1$-dimensional vector $\xbf_{1i}$ from the $d_2$-dimensional vector $\xbf_{2i}$ (with $d=d_1+d_2$).
That is, we want to design a $d_1 \times d_2$ weight matrix $\regmat$ such that $\xbf_{1i}\approx \regmat\xbf_{2i}$ for all samples $i$.
The LS approach to this supervised learning problem chooses $\hat{\regmat} = \argmin_{\regmat} \sum_{i=1}^n\|\xbf_{1i}-\regmat\xbf_{2i}\|^2$. 

Although it is possible to compute the LS solution using gradient descent, each iteration would involve the full dataset $\{\xbf_i\}_{i=1}^n$. A well-known alternative is to first build the empirical auto-correlation matrix $\hat{\Rbf} := \frac{1}{n}\sum_{i=1}^n \vx_i \vx_i\tran$ and then compute $\hat{\regmat}$ in closed-form as
\begin{align}
\hat{\regmat} = \hat{\Rbf}_{12} \hat{\Rbf}_{22}^{-1}
\text{,~with~}
\Big(\begin{smallmatrix}\hat{\Rbf}_{11} & \hat{\Rbf}_{12} \\ \hat{\Rbf}_{21} & \hat{\Rbf}_{22} \end{smallmatrix}\Big)
= \hat{\Rbf},
\label{eq:LS}
\end{align}
where each sub-matrix $\hat{\Rbf}_{ij}$ has dimension $d_i \times d_j$. Here we again use the sketch~\eqref{eq:sketch-def} with the feature map $\fmap(\Vec{x})=\vect(\Vec{x}\Vec{x}\tran)$ and extract the target parameter $\hat{\regmat}$ from~\eqref{eq:LS}.

\paragraph{Gaussian-mixture modeling}
Here the objective is to find the parameters 
$\thetabf=\{\alpha_\ell,\mubf_\ell,\Sigmabf_\ell\}_{\ell=1}^k \subset \Thetabf$ that best fit a $k$-term Gaussian-mixture model
$p(\xbf|\thetabf) = \sum_{\ell=1}^k \alpha_\ell \cN(\xbf;\mubf_\ell,\Sigmabf_\ell)$
to the data $\{\xbf_i\}_{i=1}^n$.
The parameter space $\Thetabf$ demands that, for each $\ell$: $\alpha_\ell > 0$, $\bs \mu_\ell$ is a $d$-dimensional centroid, $\Sigmabf_\ell$ is a $d \times d$ positive definite matrix, and $\sum_\ell \alpha_\ell=1$.
Traditionally, expectation-maximization (EM)~\ocite{moon1996expectation} is used to approximate
the maximum likelihood (ML) estimate $\hat{\thetabf} = \argmax_{\thetabf\in\Thetabf} \sum_{i=1}^n \log p(\xbf_i|\thetabf)$,
but the EM algorithm processes all $n$ data samples $\{\xbf_i\}_{i=1}^n$ at each iteration, which can be computationally burdensome when $n$ is very large.

\begin{highlight}[label={yb:audio}]{
{\modifiedtext Compressive \unmodifiedtext} 
Mixture Modeling for Speaker Verification}

{
\begin{center} \parbox{0.8\textwidth}{
	
  \includegraphics[width=0.8\textwidth]{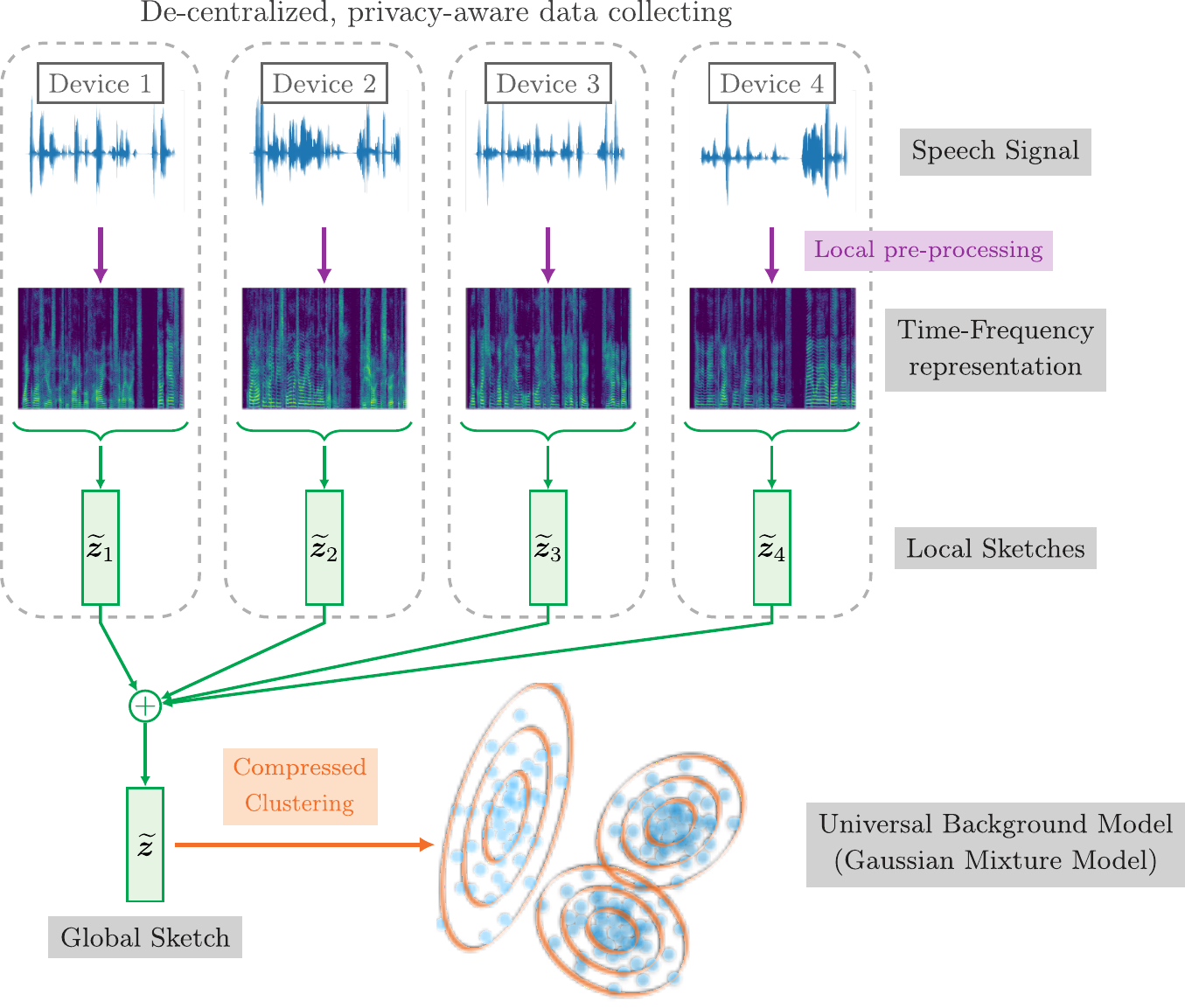}
  \captionof{figure}{\small Speaker verification \new{via compressive learning}. Unlabeled training speech data is collected in a decentralized manner, pre-processed, and locally sketched. The local sketches are then merged \new{into a global sketch, from which} a Universal Background Model is learned.}
  \label{fig:audio}
}
\end{center}
}
Given a fragment of speech and a candidate speaker, the goal of speaker verification is to assess whether the fragment was indeed spoken by that person. 
A classic approach to speaker verification is the GMM-UBM \cite{Reynolds2000} (\new{Gaussian Mixture Model}--Universal Background Model). 
\new{There, the idea is} to train a model of a ``universal'' speaker from unlabeled training data and then compare it to a specialized model for the candidate speaker. 
\new{Using the speech fragment,} the likelihood ratio between the specialized and universal models is computed, \new{and a positive decision is made if the ratio} exceeds a certain threshold.
In GMM-UBM, the data is modeled using a \new{GMM}. 
That is, $p(\xbf|\thetabf)$ are multivariate Gaussian distributions applied to a suitable time-frequency transform of the raw audio signal \cite{Reynolds2000} (see Fig.~\ref{fig:audio}).

The training of the UBM is computationally demanding, since it must be done on a large \new{corpus} of speech data. 
Moreover, the latter must be collected in a wide range of situations, so that the final model may be as universal as possible. 
For this reason, the data collecting process is best performed in a decentralized manner. 
Finally, speech data collected in real-life situations is known to be sensitive \new{information}. 
\new{For all of these reasons, compressive learning is well suited} to speaker verification.

In \cite{keriven2017b}, using an RF feature map $\fmap(\cdot)$, the authors compressed 1000 hours of speech data (50 gigabytes) into a sketch of a few kilobytes on a single laptop. 
Subsequently, they performed GMM estimation, \new{i.e., learning,} using a greedy algorithm, thereby tackling the optimization problem~\eqref{eq:learnmixture} \finalrevisions{introduced later}. 
They compared this \new{compressive}-learning approach to the EM algorithm, which, on the same laptop, could only be trained using 5 hours of speech given the available RAM. 
They observed that, by enabling the use of more data within a fixed memory budget, sketching produced better results, despite the tremendous compression factor
(see Fig.~\ref{fig:DET}).

{
\begin{center} \parbox{0.8\textwidth}{\centering
\includegraphics[width=\textwidth/3]{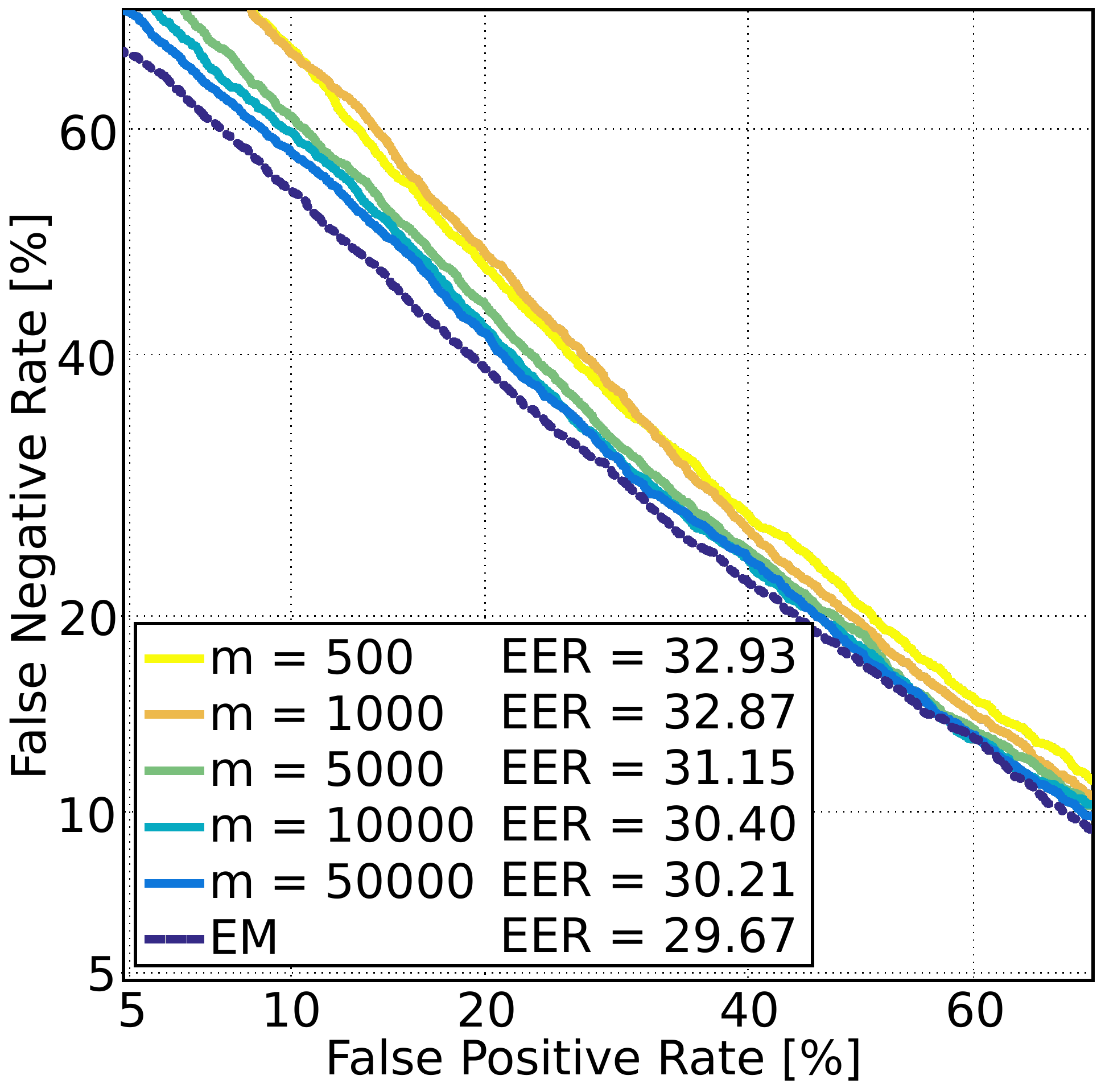}
\includegraphics[width=\textwidth/3]{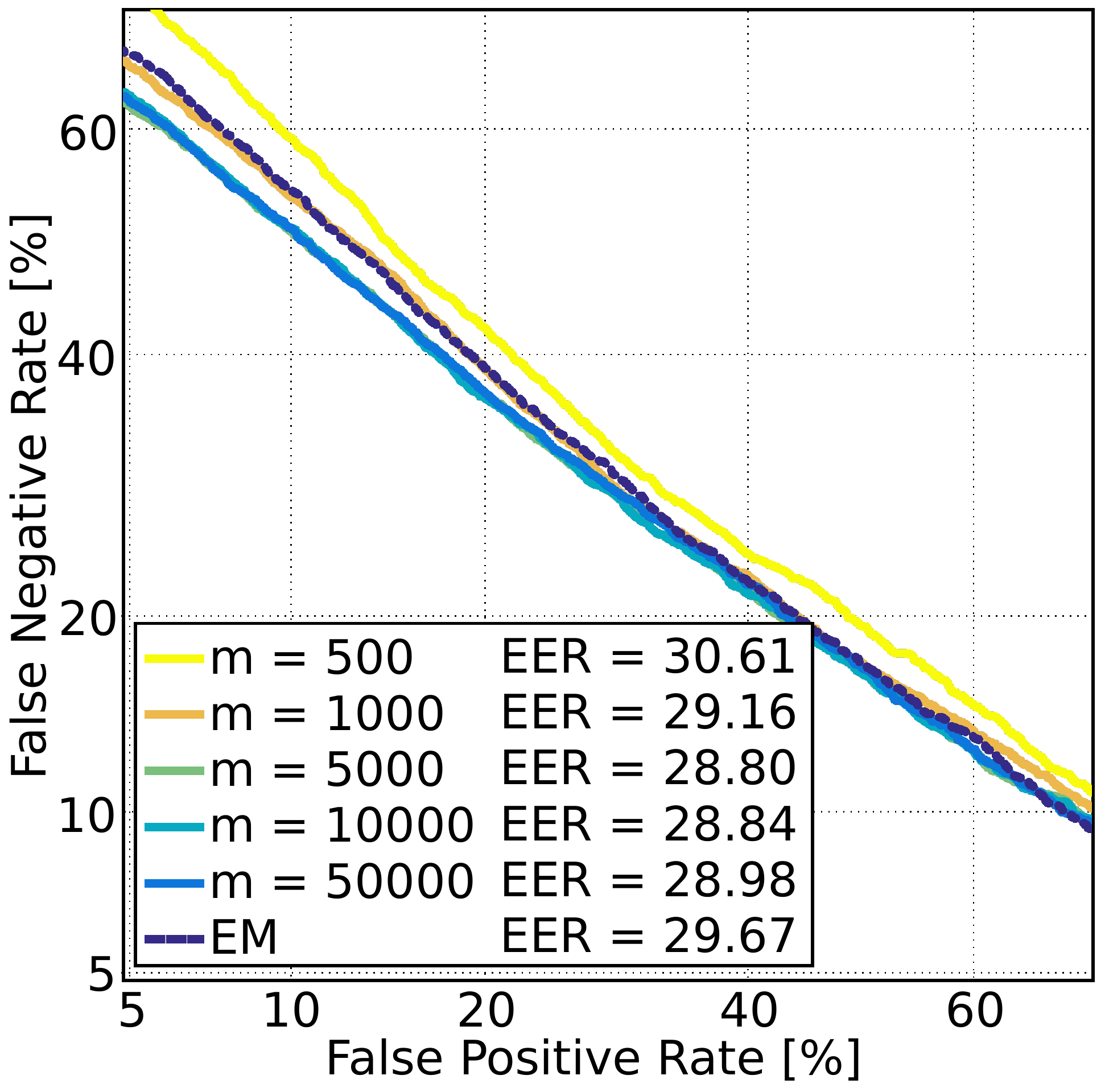}
  \captionof{figure}{\small Detection Error Tradeoff (DET) curve (False-Positive/False-Negative tradeoff obtained by varying the decision threshold) and Equal Error Rate (EER). On a single laptop: EM was trained on $5h$ of speech, \new{compressive} GMM estimation used $5h$ in the left figure and $1000h$ in the right, for different sketch dimensions \new{$m$}.}
  \label{fig:DET}
}
\end{center}
}
\end{highlight}

An alternative~\cite{keriven2017b} is to compute a sketch $\empts$ of the form~\eqref{eq:sketch-def} using random Fourier (RF) features~\cite{Rahimi2007}:
\begin{align}
\Phibf(\xbf) = \exp(-\jmath 2\pi \linop \xbf), 
\label{eq:RFF}
\end{align}
where $\jmath=\sqrt{-1}$ and $\linop\in\mathbb{R}^{m\times d}$ is a realization of a random matrix (\eg \iid Gaussian).
Here, $\exp(\cdot)$ is the componentwise exponential, not the matrix exponential.
\new{We will have a lot to say about the RF feature map later in this article.\footnote{\modifiedtext For example, the connection between the RF map and the Gaussian kernel is discussed in \ybref{yb:Mercer}.}}
The parameters can then be extracted by optimizing a cost function as explained later.
This approach has been applied to audio source-separation~\ocite{Keriven2018b} as well as speaker verification~\cite{keriven2017b}, where it was shown that $1000$ hours of speech can be compressed down to a few kilobytes without loss of verification performance (see~\ybref{yb:audio}).

\paragraph{\kmeans clustering}

The goal of clustering is to group together ``similar'' data samples from $\{\xbf_i\}_{i=1}^n$.
In the \kmeans approach to clustering, one aims to find the set of $k$ centroids $\{\cbf_\ell\}_{\ell=1}^k$ that minimizes the average squared distance from each sample to its nearest centroid, \ie $\sum_{i=1}^n \min_\ell \|\xbf_i-\cbf_\ell\|^2$. 
The famous Lloyd algorithm~\ocite{Lloyd1982}\cite{gray1998quantization} is typically used in an attempt to solve this problem.
When $n$ is very large, however, Lloyd's algorithm becomes computationally demanding.
Instead, one could sketch the dataset using~\eqref{eq:sketch-def} and extract the centroids from the sketch~\cite{keriven2017a,byrne2019sketched}. 
For this purpose, one could use RF features~\eqref{eq:RFF} and (as explained in the sequel) solve for the centroids $\{\cbf_\ell\}_{\ell=1}^k$ and the (non-negative, sum-to-one) weights $\{\alpha_\ell\}_{\ell=1}^k$ that minimize
$\|\empts - \sum_{\ell=1}^k \alpha_\ell \Phibf(\cbf_\ell)\|$.
\finalrevisions{In this setting, the weights allow us to model unbalanced clusters, yet only the centroids need to be recovered.}
On large datasets, this approach can be orders-of-magnitude better than \new{Lloyd's algorithm} in memory and runtime, provided that the sketch dimension $m$ is large enough, 
\ie that $m$ is on the order of $kd$, where $kd$ is the number of free parameters in $\{\cbf_\ell\}_{\ell=1}^k$~\cite{keriven2017a,byrne2019sketched}. 
\modifiedtext
For example, this method allows us to cluster the MNIST digit dataset, of dimension $d=784$ and cardinality $n=70\,000$, using a complex-valued sketch of dimension $m = 400$ (see \ybref{yb:MNIST}).
\unmodifiedtext

\begin{highlight}[label={yb:MNIST}]{
{\modifiedtext Compressive \unmodifiedtext} 
Clustering of MNIST digits}
{
\begin{center}
\parbox{0.8\textwidth}{\centering
\includegraphics[width=0.7\textwidth]{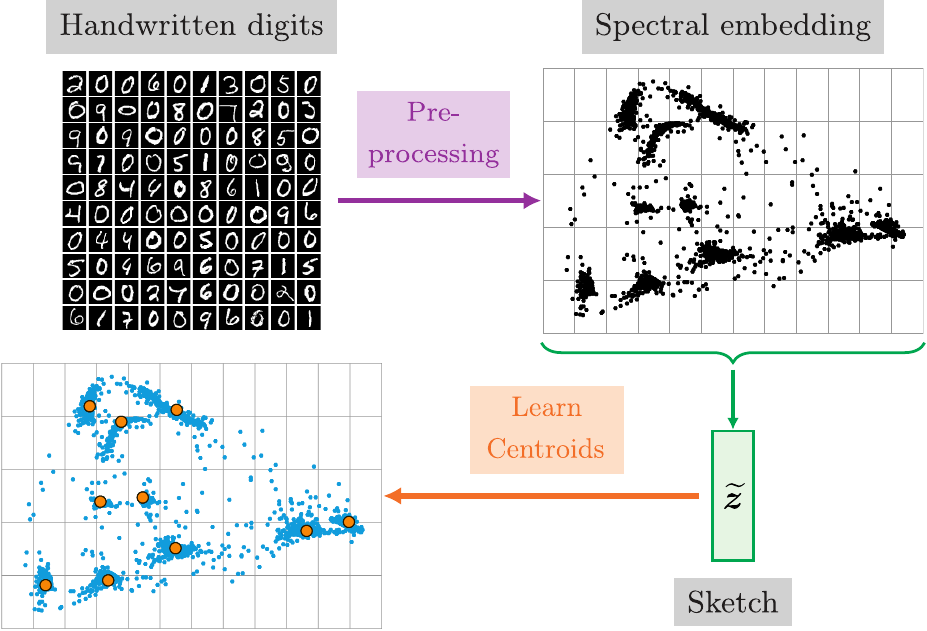}
\captionof{figure}{\small \new{Clustering of handwritten digits via compressive \kmeans. First,} SIFT descriptors are extracted from each image. Then a similarity graph is constructed 
\new{to obtain a so-called \emph{spectral embedding} of the dataset (using the first eigenvectors of the Laplacian of the graph). The spectral features are aggregated into a sketch, from which centroids are extracted.}
\label{fig:CKM_illus}
}
}
\end{center}
}

\new{Clustering is a classic machine-learning task and a component of many machine-learning} pipelines. 
The most popular method is Lloyd's algorithm \cite{gray1998quantization}\ocite{Lloyd1982}, \new{which aims to solve} the \kmeans problem. 
\new{In many cases, the raw features are converted to a spectral embedding before clustering.}
As a proof of concept, the authors in \cite{keriven2017a} applied this technique to handwritten digits from the MNIST dataset, but used \new{compressive \kmeans} clustering \new{in place of Lloyd's algorithm (see Fig.~\ref{fig:CKM_illus}). 
Using a $n=10^7$-sample augmentation of MNIST,} they found that compressive \new{\kmeans} clustering gave approximately the same accuracy as \new{Lloyd's algorithm} but reduced the time- and memory-complexity by \new{1.5} and 4 orders-of-magnitude, respectively (see Fig.~\ref{fig:CKM}).

{\begin{center}
\parbox{0.8\textwidth}{\centering
\includegraphics[width=0.3\textwidth]{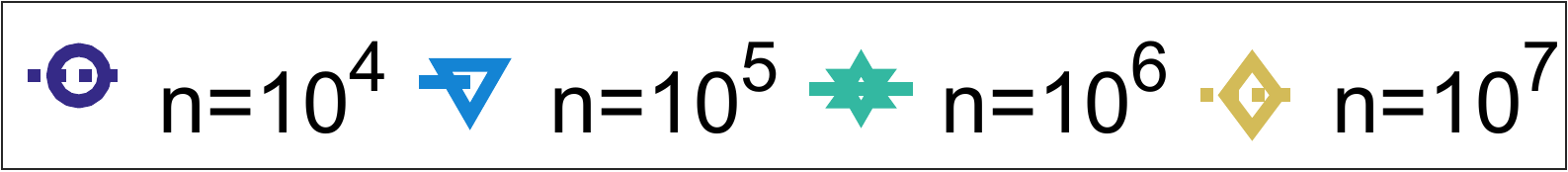}\\
\includegraphics[width=0.6\textwidth]{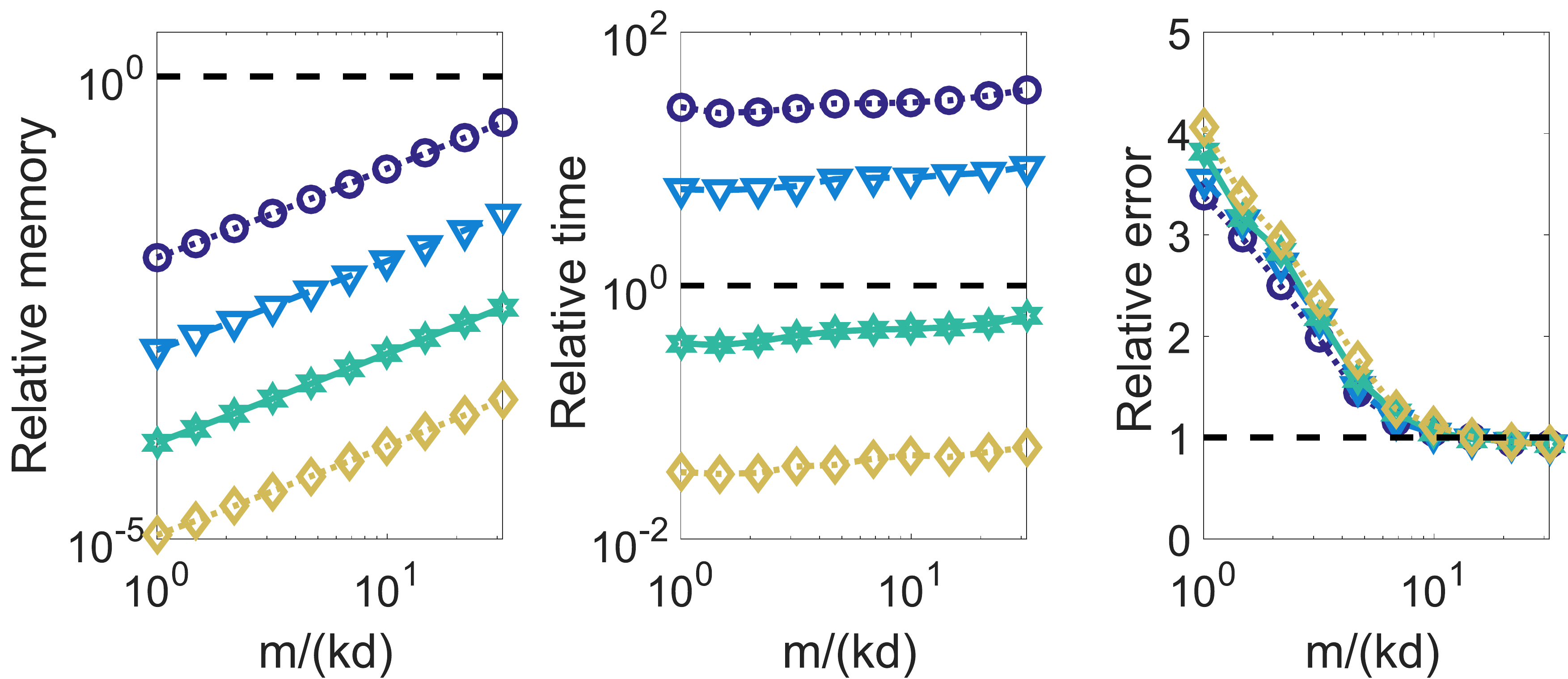}
\captionof{figure}{\small Memory, runtime, and error of \new{compressive \kmeans} clustering relative to Lloyd's algorithm \new{for various sample cardinalities $n$ and sketch lengths $m$, using $k=10$ centroids and $d=10$ dimensional spectral features}. Figure from \cite{keriven2017a}.}
\label{fig:CKM}
}\end{center}
}

\end{highlight}

Although this paper focuses on these examples, the general 
\modifiedtext compressive\unmodifiedtext-learning framework extends beyond them. 
It has been applied to, for example, Independent Component Analysis~\cite{sheehan2019compressiveICA} (by sketching the cumulant tensor) and subspace clustering~\cite{sheehan2019compressive} (using a polynomial feature map). 
Extending the framework to a new task raises essential questions such as: How do we choose the feature map $\Phibf(\cdot)$?
How do we learn the essential parameters $\thetabf$ from the sketch $\empts$? Are there statistical learning guarantees?
Can sketching and learning be made computationally efficient? 
Can we sketch while respecting privacy? 
This paper sketches answers to these questions for the worked examples introduced above.

\begin{highlight}[label={yb:conv-table}]{Notations and Conventions}
  \centering
  \scalebox{0.85}{
    \definecolor{tableShade}{gray}{0.95}
    \rowcolors{2}{tableShade}{highlightbg}
    \begin{tabularx}{1.02\linewidth}{ll}
      \hiderowcolors 
      {\bf Notation}&{\bf Description}\\
      \showrowcolors 
      \toprule
      $\mathcal X = \{\vx_i\}_{i=1}^n$ & dataset of $n$ training samples $\vx_i \in \mathbb R^d$\\ 
      $\empts$ & (empirical) sketch of the dataset, a vector of dimension $m$\\
      $\fmap(\cdot)$ & sketching feature map, a function mapping $\mathbb R^d$ to either $\mathbb R^m$ or $\mathbb C^m$\\ 
      $\thetabf \in \Thetabf$ & target parameters to learn, of dimension $p$ (\eg PCA matrix, centroids, mixture model)\\ 
      $\linop$ & random matrix associated with certain feature maps 
      $\fmap(\cdot)$, usually drawn \iid Gaussian\\ 
      $\varrho(\cdot)$ & componentwise nonlinearity associated with certain feature maps 
      $\fmap(\cdot)$\\ 
      \whp& with high probability, \ie with exponentially decaying failure probability\\
      \iid&independent and identically distributed\\
      \wpone&with probability 1\\
  $\|\cdot\|$,
      $\langle \cdot,\cdot\rangle$ 
      & Euclidean norm of a vector, inner product between vectors\\      $\|\cdot\|_{0}$ & $\ell^{0}$ norm of a vector, \ie its number of nonzero entries\\
      $\|\cdot\|_{F}$ & Frobenius norm of a matrix\\
     $\|\cdot\|_{L^{2}}$,   $\langle \cdot,\cdot\rangle_{L^{2}}$ & $L^{2}$ norm of a function, $L^{2}$ inner product between functions      
    \end{tabularx}
  }
\end{highlight}

\section{Historical background}

The term ``sketch'' has different meanings, depending on the field. 
Our use of the term comes from the literature on relational databases, and more particularly from the subfield of \emph{Approximate Query Processing} (AQP)~\cite{Cormode:2012gh}. 
The goal of AQP is to build a short description of the content of a massive dataset, called a {\em synopsis}, by analogy with the synopsis of a movie or a book, such that certain \emph{queries} can be efficiently performed to return answers with controlled error and/or probability of failure. 
Well-known queries include the frequency of occurrence of a particular element in a stream of data (taken from a discrete collection), and the minimum of several of these frequencies, yielding the celebrated count-min sketch~\cite[Section 5.3.1]{Cormode:2012gh}\ocite{Cormode2005cm} synopsis. 
In this context, the statistical parameters $\thetabf$ learned from a sketch are interpreted as the result of a particular query on the dataset. 

Despite the apparent similarities between AQP and 
\modifiedtext compressive \unmodifiedtext 
learning, both datatypes and learning tasks differ significantly between the two fields.
Typically, AQP focuses on (multi)sets of elements taken from a discrete collection of objects and considers database queries and related operations, while 
\modifiedtext compressive \unmodifiedtext
learning focuses on continuous-valued signals (\eg images or audio signals) and considers machine-learning tasks such as density estimation or regression.

\subsection{
\modifiedtext Sketches \unmodifiedtext for streaming and distributed methods}
\begin{figure}[t]
        \begin{center}
            \ifrecompiletikz
				\input{fig-distributed_streaming}
			\else
				\includegraphics[width=.9\linewidth]{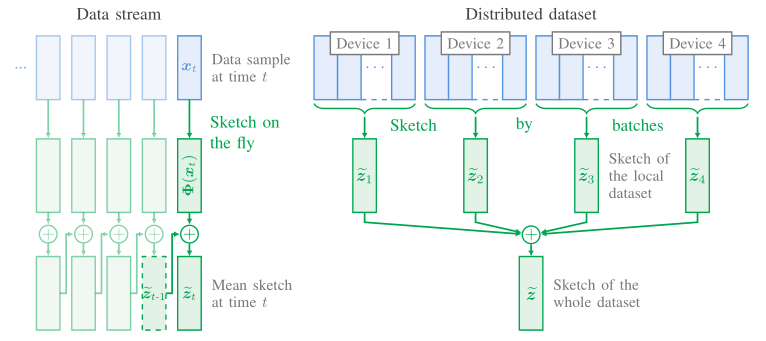}
			\fi
        \end{center}
        \caption{\small Left: A streaming scenario, where the data samples are sketched one-by-one, and the mean sketch is updated at each time. Right: A distributed scenario, where each device computes a local sketch, and a centralized entity further averages these local sketches.}
        \label{fig:distributedstreams}
  \end{figure}
      
In AQP, a popular class of synopses is that of \emph{linear sketches}~\cite[Chap. 5]{Cormode:2012gh}.  
A linear sketch, \new{which maps a dataset to} a vector, must satisfy the single condition that \emph{the sketch of the concatenation of two datasets \new{equals} the sum of their sketches}. 
In mathematical terms, if we denote by $\empts(\ds)$ the sketch of a dataset $\ds = \{\vx_i\}_{i=1}^n$, then it is required that $\empts(\ds) = \empts(\ds_1) + \empts(\ds_2)$ whenever $\ds$ is the concatenation of $\ds_1$ and $\ds_2$.  
It thus follows that a linear sketch must be of the form $\empts\prt{\ds} = \sum_{i=1}^n \fmap(\vx_i)$ for some \new{possibly nonlinear} feature map $\fmap(\cdot)$. 
That is the same definition that we adopt in~\eqref{eq:sketch-def}, except that we normalize by the number of elements $n$.
Linear sketches are popular in AQP mainly because they are well suited to \emph{streaming} scenarios. 
That is, inserting or deleting an element $\vx_i$ from the dataset corresponds to adding or subtracting $\fmap(\vx_i)$ from the sketch, respectively (see Fig.~\ref{fig:distributedstreams}). 
Note that some very simple sketches are not linear: for instance, it is easy to sketch the maximum running value in a stream of scalar data by computing the maximum of the current sketch and each new data point, but this sketching procedure is not linear \cite[Chap. 5.2.1]{Cormode:2012gh}.
In particular, this ``max'' sketch facilitates the insertion of a new element in the database, but not the deletion of an existing one.

\modifiedtext
\begin{remark*}
In the subsections that follow, we describe linear sketches from other points of view. 
Before doing that, however, we want to clarify that the terms ``sketching'' and ``linear sketching'' appear in many other fields, although with meanings that differ considerably from ours. 
For instance, sketching is often used to indicate \textbf{linear} dimensionality reduction in $n$ or $d$, as attained by multiplying the data matrix $[\vx_1, \ldots, \vx_n]^\top \in \mathbb R^{n\times d}$ by a random matrix on the left or the right \cite{Achlioptas2003, Mahoney2010, Boutsidis2010, Woodruff2014}. 
Sketching may also refer to the use of sampling-based approaches \cite{Achlioptas2002}, such as coresets \cite{Feldman2010} or the Nystr\"om method \cite{Williams2001a}.
These latter methods differ from linear sketches in AQP, which is our notion of sketching, in that these latter methods generally do not respect the concatenation condition described earlier and are therefore less directly amenable to streaming scenarios. 
Moreover, these methods typically do not involve a \textbf{nonlinear} feature map $\fmap(\cdot)$, which is a key component of our sketch.
\end{remark*}
\unmodifiedtext

\subsection{Sketches as (randomized) generalized moments}

In signal processing and machine learning, the data samples $\vx_i$ generally live in the vector space $\bR^{d}$
and are often modeled as independent and identically distributed (\iid) random vectors having a \emph{probability distribution} with density $p_X$. 
Consider what happens when the number of samples $n$ goes to infinity. 
The strong law of large numbers says that 
\begin{equation}\label{eq:asymptoticsketchasmoment}
\lim_{n\rightarrow\infty} \frac{1}{n}\sum_{i=1}^n \Phibf(\vx_i) \stackrel{\wpone}{=} \Exp[\fmap(X)] =\int p_X(\xbf) \fmap(\vx) \dif\vx,
\end{equation}
where $\Exp[\cdot]$ denotes expectation with respect to the probability density $p_X$. 
If we consider the simple case of dimension $d=1$ and the scalar transformation $\fmap(x) = x^k$ (so that $m=1$), then $\Exp[\fmap(X)]$ is the (uncentered) $k$-th \emph{moment} of the random variable $X$, a quantity that has a long history in statistics. 
By analogy, with a generic vector-valued feature map $\fmap(\cdot)$ and in dimension $d>1$, quantities of the form $\Exp[\fmap(X)]$ are known as \emph{generalized moments} of the 
random vector $X \in \mathbb{R}^{d}$.

Performing inference from generalized moments is often referred to as the \emph{Generalized Method of Moments} (GeMM) \cite{hall2005}.
This method \modifiedtext to ``learn from a sketch'' \unmodifiedtext  is very popular in, \eg the field of econometrics \cite[Chap. 1]{hall2005}.
The GeMM can be seen as an alternative to maximum likelihood  
estimation that avoids the need to work with the full likelihood function, which can have computational benefits.
Indeed, for many classes of probability distributions, such as heavy-tailed $\alpha$-stable distributions, the likelihood function is not given in closed form, but generalized moments are given in closed form for appropriately chosen feature maps $\fmap(\cdot)$ (see~\ybref{yb:audio}).

GeMM differs from 
\modifiedtext compressive \unmodifiedtext 
learning in several aspects.
In GeMM, the feature map $\fmap(\cdot)$ is typically constructed to make the parameter estimates computable in closed form.
This narrows the range of learning tasks that GeMM can handle.
In 
\modifiedtext compressive \unmodifiedtext 
 learning, $\fmap(\cdot)$ is designed with information-preservation in mind.
Consequently, the range of learning tasks is much broader in 
\modifiedtext compressive \unmodifiedtext 
learning, although the estimation procedure \modifiedtext (i.e. \emph{learning from a sketch}) \unmodifiedtext may be algorithmic in nature. 
Another difference is that, in 
\modifiedtext compressive \unmodifiedtext 
learning, $\fmap(\cdot)$ is typically randomized.
This results in 
\emph{randomized generalized moments}, 
which are rarely seen in 
GeMM.

\subsection{\modifiedtext Compressive \unmodifiedtext learning and compressive sensing}

\modifiedtext As we will show in this section, the \unmodifiedtext
sketching mechanism~\eqref{eq:sketch-def} can be interpreted as a dimensionality-reducing linear ``projection'' \emph{of the probability distribution underlying the data set $\{\vx_i\}_{i=1}^n$}. This differs from the traditional approach in signal processing, where dimensionality reduction is performed on the features $\xbf_i\in\bR^d$ themselves, rather than the distribution that generates them. 
Still, many of the intuitions, analyses, and tools designed for feature-based dimensionality reduction can be extended to distribution-based dimensionality reduction.
\ifarxiv Some details are now provided. \else \fi

In the field of \emph{inverse problems} and \emph{compressive sensing} (CS)~\cite{candes2008introduction,FouRau13}, 
a signal-of-interest is modeled as a high-dimensional vector $\vx \in \bR^d$, and a physical measurement of that signal is approximated by a linear transformation plus additive noise: $\Vec{y} = \Vec{A} \vx + \noise\in \bR^m$ (see~\ybref{yb:CS}).
Here, the linear measurement operator 
is represented by the $m\times d$ matrix $\Vec{A}$.
In many applications, there is a great motivation to make the measurement dimension much less than the signal dimension, \ie $m\ll d$.
In this case, the recovery of $\vx$ from $\Vec{y}$ is generally ill-posed.
Still, it is possible to accurately recover $\vx$ from $\Vec{y}$ when both $\vx$ and $\Vec{A}$ obey certain properties and the noise is small enough. 

\modifiedtext The celebrated Johnson Lindenstrauss (JL) lemma and its variations \ocite{johnson1984extensions,Achlioptas2003} specifies, for instance, that with high probability one can nearly preserve the pairwise distances between $N$ features in $\bb R^d$ by linearly projecting them in a $O(\log N)$-dimensional domain. 
This is used, for instance, to accelerate nearest-neighbor searches in large databases. 
An important extension is to compressive sensing: \unmodifiedtext
if $\vx$ is \emph{sparse}, in that relatively few of its coefficients deviate significantly from zero, and if $\Vec{A}$ satisfies the restricted isometry property (RIP)\modifiedtext---an extension of the JL lemma to the continuous set of sparse signals---\unmodifiedtext then one can pose a \emph{regularized} inverse problem whose solution is close to the true $\vx$ (see~\ybref{yb:CS}).

\begin{highlight}[label={yb:CS}]{Compressive Sensing and the Restricted Isometry Property}
In compressive sensing, one observes a linear measurement $\Vec{y} = \Vec{Ax}\in\bR^m$ of signal $\Vec{x}\in\bR^d$ with $m\ll d$, \ie with significantly reduced dimension.
(For simplicity, we focus on the noiseless case for now.)
To distinguish between different signals $\Vec{x}$ in a given signal class, one desires that the distances between all signals in that class are preserved by the measurement operator $\Vec{A}$. 
For the class of $k$-sparse signals $\Sigma_k$, \ie signals with at most $k$ non-zero entries, this property is satisfied up to a tolerance $\delta\in[0,1)$ 
when $\Vec{A}$ obeys the restricted isometry property (RIP) \cite{candes2008introduction,FouRau13}\ocite{Candes2004decodlinear}
\begin{equation}
(1 - \delta) \|\bs x - \bs x'\|^2 \leq \|\bs A \bs x - \bs A \bs x'\|^2 \leq (1 + \delta) \|\bs x - \bs x'\|^2,\ \forall \bs x,\bs x' \in \Sigma_k 
\label{eq:RIP}.
\end{equation}
One way to create a RIP-satisfying $\Vec{A}$ is to draw it randomly.
For example, if the coefficients of $\Vec{A}$ are drawn \iid zero-mean Gaussian, then $\Vec{A}$ will satisfy the RIP \emph{with high probability} (\whp) when 
the sketch dimension $m$ is at least on the order of $k \log(d/k)$
\cite{FouRau13}.

To recover $k$-sparse $\Vec{x}$ from $\Vec{y}$, one might attempt to search for the sparsest signal among all of those that agree with the measurements, \ie
within the set $\cl C_{\bs y} := \{\ubf : \bs A \ubf = \bs y\}$.
The complexity of this search, however, grows exponentially in $k$.
Fortunately, when $\Vec{A}$ satisfies the RIP, one can provably recover the true $\Vec{x}$ using polynomial-complexity methods \cite{FouRau13}.
One approach is to solve the convex problem of finding the signal with the smallest $\ell_1$-norm within $\cl C_{\bs y}$.
Another is to use a greedy algorithm, like orthogonal matching pursuit (OMP), which estimates $\bs x$ by progressively removing from $\bs y$ the $k$ columns of $\bs A$ that best ``align'' with it, using a least-squares fit.

The RIP also provides guarantees on robust recovery.
Suppose that we have noisy measurements $\ybf =  \Abf \xbf+\noise$, with noise $\noise$ of bounded norm $\|\noise \| \leq \varepsilon$.
In addition, suppose that $\xbf$ is only \emph{approximately} $k$-sparse, and use $\xbf_k$ to denote the best $k$-sparse approximation of $\xbf$.
Finally, consider recovering an estimate $\widehat{\xbf}$ of $\xbf$ by searching for
the signal with smallest $\ell_1$-norm 
that agrees with the measurements up to a tolerance of $\varepsilon$, \ie within the set $\cl C_{\bs y, \varepsilon} := \{\ubf : \|\bs A \ubf - \bs y\| \leq \varepsilon \}$.
Then, if the RIP~\eqref{eq:RIP} holds, the estimation error $\widehat{\xbf}-\xbf$ satisfies \cite{FouRau13}\ocite{candes2006robust}
\begin{equation}
  \label{eq:l1-l2-instance-optimality}
  \|\widehat{\xbf} - \xbf\| \leq C \frac{\|\xbf - \xbf_k\|_1}{\sqrt{k}} + D \varepsilon,
\end{equation}
where constants $C,D>0$ depend only on the value of $\delta$ that appears in~\eqref{eq:RIP}. 
Thus, the estimation error increases linearly with the noise level $\varepsilon$ and the deviation $\|\xbf-\xbf_k\|_1$ from perfect sparsity.
  
\centering \parbox{0.8\textwidth}{
  \medskip
  \includegraphics[width=0.8\textwidth]{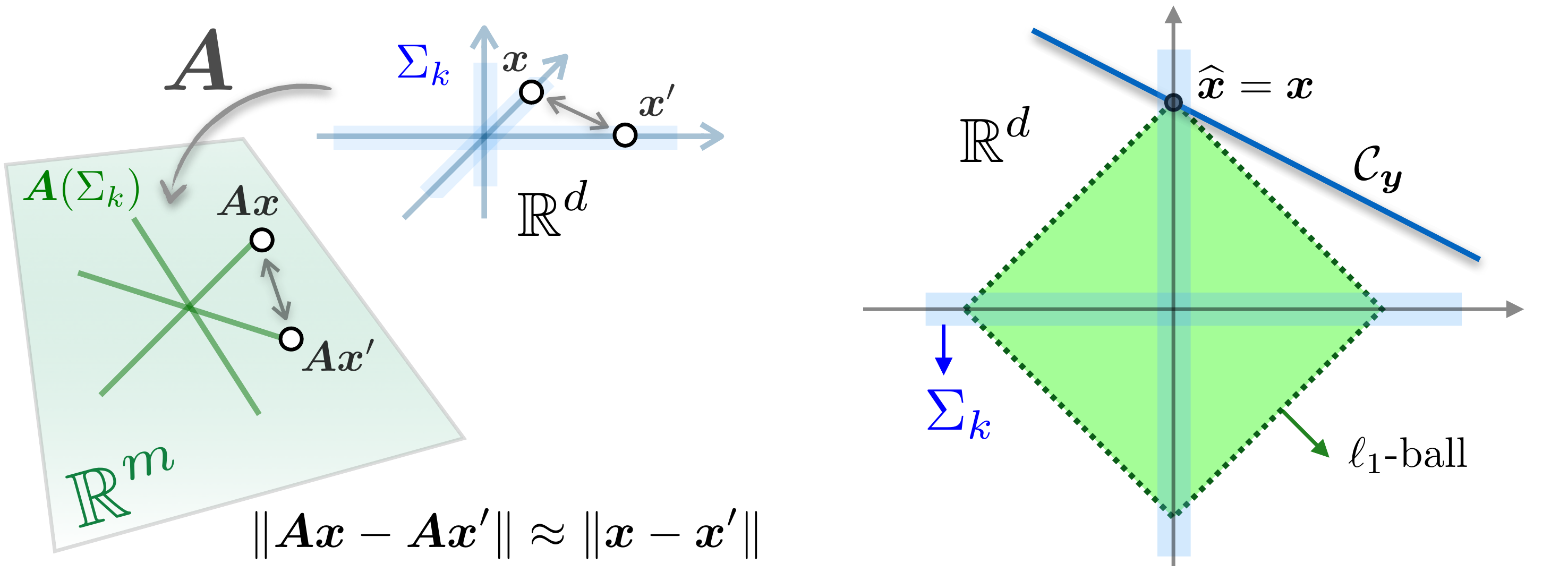}
  \captionof{figure}{\small (left) Geometrical interpretation of the RIP. (right) 2-D illustration of the recoverability of $\vx$ from $\vy = \bs A \bs x$ by finding the vector $\widehat{\xbf}$ in $\cl C_{\vy}$ with smallest $\ell_1$-norm.}
  \label{fig:spm-cs}
}
\end{highlight}

Now that CS has been described, we can clearly connect it to 
\modifiedtext compressive \unmodifiedtext 
learning.
Recall that the sketch~\eqref{eq:sketch-def} $\empts$ converges to the generalized moment $\ts := \bE[\fmap(X)] = \int p_X(\vx) \fmap(\vx) \dif\vx$ as the number of data samples $n$ tends to infinity, as per~\eqref{eq:asymptoticsketchasmoment}.

A crucial observation is the following: \emph{due to the linearity of integration, the sketch $\ts$ depends linearly on the probability density $p_X$.}
To see it another way, consider a mixture of two densities, $p_{X} = \alpha p_{X_{1}}+(1-\alpha)p_{X_{2}}$. 
The corresponding expectation satisfies
\begin{equation}\label{eq:sklinearity}
\bE_{X \sim p_{X}}[\fmap(X)] = \alpha \bE_{X_{1} \sim p_{X_{1}}} [\fmap(X_{1})] + (1-\alpha) \bE_{X_{2} \sim p_{X_{2}}} [\fmap(X_{2})],
\end{equation}
which implies that the generalized moment $\ts$ is linear in $p_X$.
With this understanding, we can write 
\begin{equation}\label{eq:ProjDef}
\ts = \cA(p_{X}) := \bE_{X \sim p_{X}}[\fmap(X)] ,
\end{equation}
where $\cA$ is a linear operator mapping the probability distribution $p_X$ to the $m$-dimensional sketch vector~$\ts$.
\begin{remark*}
Although $\cA$ is a linear function of $p_X$, we emphasize that the feature map $\fmap(\vx)$ is generally \emph{not} a linear function of $\vx$. \modifiedtext This makes compressive learning concretely very different from the vast majority of existing ``sketching'' mechanisms, which use random {\em linear} projections of the data for dimensionality reduction. \unmodifiedtext
\end{remark*}

\begin{figure}[t]
\centering
\includegraphics[width=0.7\textwidth]{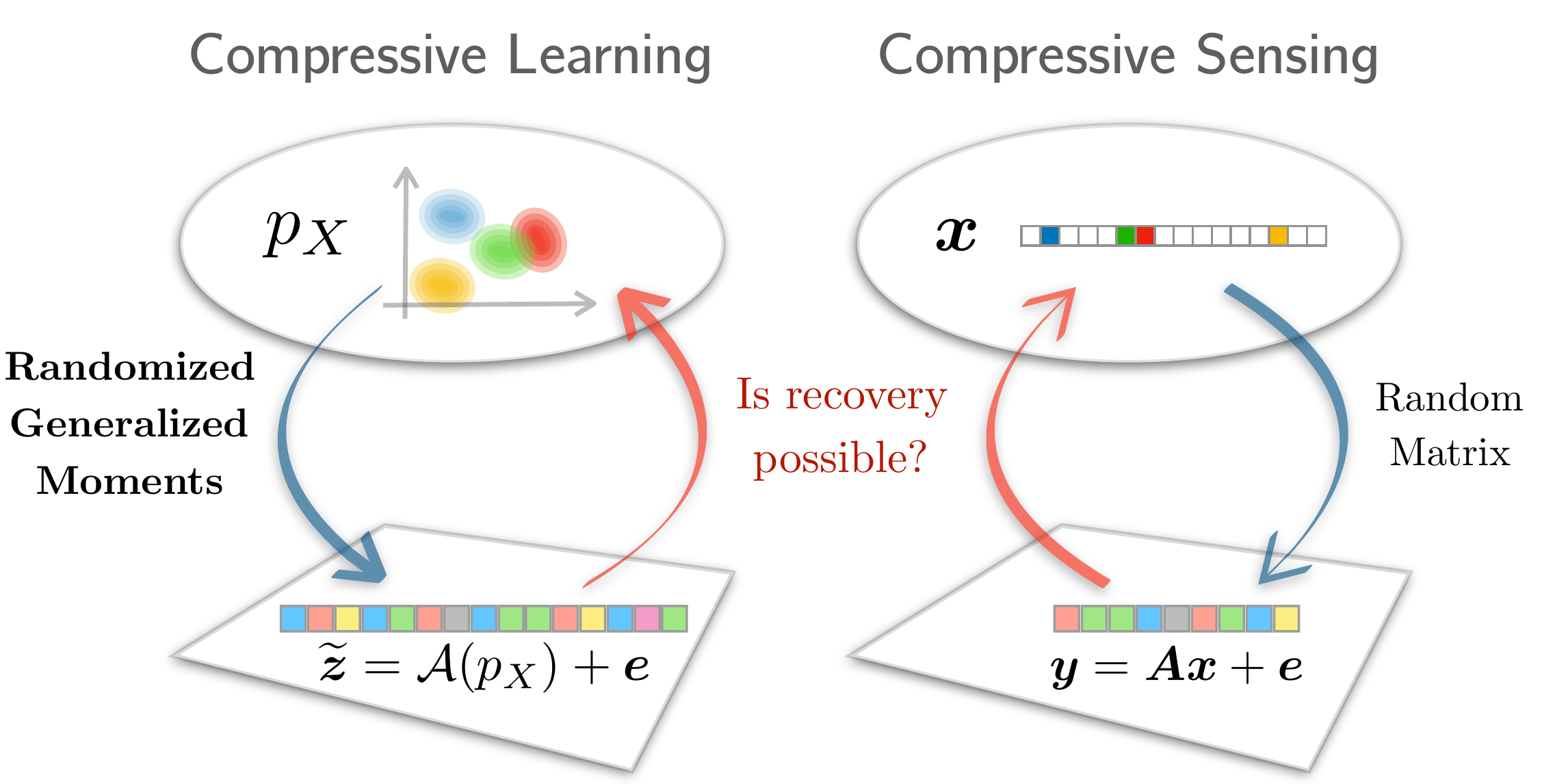}
\caption{\small 
The analogy between \modifiedtext compressive \unmodifiedtext
learning, which uses the dimensionality reducing linear measurement $\cA(p_X) = \bE[\fmap(X)]$ of a distribution $p_X$, and compressive sensing, which uses the dimensionality reducing linear measurement  $\bs A \bs x$ of a signal $\vx$.}
\label{fig:skvscs}
\end{figure}

With a small modification of the above arguments, we can handle the finite-sample case.
Consider the difference between the true generalized moment $\ts$ and the empirical moment $\empts$, \ie
\begin{equation}\label{eq:noise}
\frac{1}{n} \sum_{i=1}^n \fmap(\vx_i) - \bE[\fmap(X)] =: \noise .
\end{equation}
As we discussed earlier, $\noise$ converges to zero as $n$ tends to infinity by the law of large numbers. 
Combining~\eqref{eq:sketch-def},~\eqref{eq:ProjDef}, and~\eqref{eq:noise}, we obtain
\begin{equation}\label{eq:sketch-linear-projection}
\empts = \cA (p_X) + \noise ,
\end{equation}
which shows that the sketch~\eqref{eq:sketch-def} can be interpreted as \emph{a ``noisy'' observation of the data distribution $p_X$ through the linear measurement operator $\cA$.}
Under mild conditions\footnote{From the assumed independence of the data samples, this happens for instance if $\bb P[\|\fmap(X)\| \geq t]$ decays exponentially fast when $t$ increases.}, the central limit theorem can be used to show that 
$\norm{\noise}$ decays as $1/\sqrt{n}$
 \whp \cite[Chapter 8]{FouRau13}. 
With the above interpretation of 
\modifiedtext compressive \unmodifiedtext 
learning, one recovers all of the traditional ingredients of CS:
\begin{itemize}
\item 
The measurement operator $\cA$ is linear.
\item
The measurements $\empts$ are drastically dimension-reduced. 
In mathematical terms, probability distributions $p_X$ belong to the infinite-dimensional vector space of so-called \emph{finite} measures \cite{Candes2014}\ocite{Bredies2013}.
The operator $\cA$ maps these infinite-dimensional objects to vectors of finite dimension $m$. 
\item 
The measurement operator $\cA$ is typically designed using randomness.
This is accomplished by choosing an appropriate randomized feature map $\fmap(\cdot)$, such as one based on RF features, as in~\eqref{eq:RFF}. 
\item 
The measurements $\empts$ are noisy, as per~\eqref{eq:sketch-linear-projection}. 
\end{itemize}
The analogy between 
\modifiedtext compressive \unmodifiedtext
learning and CS is illustrated in Fig.~\ref{fig:skvscs}. 
\modifiedtext
The CS analog to the \emph{sketching phase}~\eqref{eq:sketch-def} is the \emph{signal sensing phase}, where the signal $\vx$ is \emph{linearly} mapped to the observation vector $\vy$. 
The analog to \emph{sparse recovery}, where an estimate of $\vx$ is computed from the observation $\vy$ by solving an optimization problem, is to {\em learn from a sketch}, where an estimate of the data distribution $p_X$ (or of distributional parameters $\thetabf$ of interest) is computed from the sketch $\empts$. 
Its expression as an optimization problem will be further discussed in Section~\ref{sec:learningfromsketch}.
\unmodifiedtext

\subsection{Random Fourier sampling and super-resolution recovery}\label{sec:RFsampling}

In this section, we consider the specific case of 
\modifiedtext compressive \unmodifiedtext 
clustering, which allows us to forge a concrete connection between CS and 
\modifiedtext compressive \unmodifiedtext
learning using random Fourier sampling and super-resolution recovery.

We begin by considering the goal of recovering $k$ centroids $\{\vc_{\ell}\}_{\ell=1}^{k}$ in $\mathbb{R}^d$ 
\new{with \finalrevisions{Euclidean} norm $\leq r$ that are separated from each other \finalrevisions{in Euclidean distance} by $\geq \varepsilon$}. 
A naive approach would be to discretize the $d$-dimensional cube of side-length $2r$ with a grid spacing of $\epsilon$, leading to $N=(2r/\epsilon)^{d}$ bins. 
A valid sketch of the data $\ds$ is obtained by simply computing the histogram $\hat{\pbf} \in \mathbb{R}_{+}^{N}$ over these bins (\ie by using the ``binning'' feature map).
However, the dimension of this sketch, $N$, grows exponentially in the feature dimension, $d$.
To construct a smaller sketch, one might reason that, if the data clusters tightly around $k$ points, then $\hat{\pbf}$ is close to a $k$-sparse vector. 
In this case, ideas from CS can be directly exploited.
In particular, one could use a sketched histogram \cite{Cormode:2012gh}\ocite{Thaper:2002kf} of the form $\empts =  \Abf \hat{\pbf}$, where matrix $\Abf\in\mathbb{R}^{m\times N}$ is randomly drawn with \iid Gaussian components. 
\modifiedtext Here, ``learning from a sketch'' means recovering the centroids. 
For this, \unmodifiedtext
one would first search for the best non-negative, $k$-sparse, sum-to-one vector using
\begin{equation}
\label{eq:KmeansBinned}
\tilde{\pbf}=\argmin_{\pbf\in\Sigma_k^{+}} \|\empts-\Abf\pbf\|^{2},
\end{equation} (or a convex or greedy relaxation of this problem) where $\Sigma_{k}^{+}$ here denotes the set of $k$-sparse, non-negative, sum-to-one vectors. Then, one would identify the $k$ grid locations in $\mathbb{R}^d$ corresponding to the non-zero indices of $\tilde{\pbf}$.
CS theory \cite{FouRau13} says that the support of $\tilde{\pbf}$ will be accurate for sketch dimensions 
$m$ at least on the order of $k \log N$, \ie at least on the order of $kd \log(r/\epsilon)$
Although this latter approach substantially reduces the dimension of the sketch, 
practical challenges remain when the feature dimension $d$ is large.
For example, the number of columns, $N$, in the compression matrix $\Abf$ grows exponentially in $d$, making storage and multiplication by $\Abf$ impractical.

An alternative approach could be to construct $\Abf$ using $m$ rows of the ($d$-dimensional, in this case) discrete Fourier transform (DFT) matrix, \ie by sampling the DFT at $m$ ($d$-dimensional) frequencies.
In this case, multiplication-by-$\Abf$ could be implemented by the fast Fourier transform (FFT) algorithm, and thus the matrix $\Abf$ would not need to be explicitly stored. 
Fourier-domain sampling is a familiar operation in the context of signal processing, as it forms the cornerstone for radar, medical imaging, and radio interferometry, see \eg \cite{lustig2008compressed,fessler2019optimization}
\ocite{Wiaux_2009,cetin2014sparsity,greco2019advances}.
When the $m$ frequencies are drawn uniformly at random, CS theory \cite{candes2008introduction,FouRau13}\ocite{candes2006robust} has established that accurate recovery of an $N$-length $k$-sparse signal can be accomplished (with high probability) when $m$ is on the order of 
$k \log^{3}(k) \log N$ \cite[Corollary 12.38]{FouRau13}.
Since $N = (2r/\epsilon)^{d}$ here, this would mandate sketch dimensions $m$ at least on the order of 
$kd \log^{3}(k) \log (r/\epsilon)$.
Although the FFT avoids the need to store $\Abf$ as an explicit matrix
and allows efficient computation of the sketch, 
the cost of
solving the optimization problem \eqref{eq:KmeansBinned} 
using existing convex relaxations or greedy approaches is impractical due to the need to manipulate $N$-dimensional vectors, where $N$ grows exponentially in $d$.

Until now, we considered discretizing the $d$-dimensional feature space on an $\epsilon$-spaced hypergrid within a $2r$-sidelength hypercube, but found that this requires manipulating vectors (\eg a histogram $\hat{\pbf}$)
whose dimension grows exponentially in $d$.
We can avoid this discretization
(\ie take $\epsilon\rightarrow 0$ and $r\rightarrow\infty$) by replacing the DFT with the \emph{continuous} Fourier transform, in which case we are Fourier-transforming the empirical distribution $\hat{p}_{\ds}(\vx) :=\frac{1}{n}\sum_{i=1}^n \delta(\vx-\vx_i)$ rather than its $N$-bin histogram $\hat{\pbf}$.
The sketch $\empts$ then has components 
\begin{align}
\tilde{z}_j 
&= \int_{\mathbb{R}^d} \hat{p}_{\ds}(\vx) \exp(-\jmath 2\pi \vw_j\tran \vx) \dif\vx
\label{eq:RFFsketch} \\
&= \frac{1}{n}\sum_{i=1}^n \exp(-\jmath 2\pi \vw_j\tran \vx_i) 
\qquad j=1,\dots,m,
\label{eq:RFFsketch1}
\end{align}
where $\{\vw_j\}_{j=1}^m$ are $d$-dimensional frequencies that are drawn at random. 
\modifiedtext 
Typically, $\vw_j$ are drawn i.i.d.~Gaussian, but other distributions can be used, as discussed in Sections~\ref{sec:design}~and~\ref{sec:matrices}.
\unmodifiedtext
Note that~\eqref{eq:RFFsketch1} corresponds precisely to the sketch~\eqref{eq:sketch-def} with the random Fourier (RF) feature map $\fmap(\cdot)$ from~\eqref{eq:RFF}. 
Taking a statistical perspective, the components $\tilde{z}_j$ in~\eqref{eq:RFFsketch} can also be recognized as samples of the \emph{characteristic function} of $\hat{p}_{\ds}$.
Recall that, for a density $p$, the characteristic function $\Psi_p$ is defined as 
\ifarxiv (the complex conjugate of) its Fourier transform, \ie 
\else \fi 
\begin{equation}
\Psi_p(\vw):=\int_{\mathbb{R}^d} p(\vx) \exp(\jmath 2\pi \vw\tran\vx)\dif\vx = \bE_{X\sim p} [\exp(\jmath 2\pi \vw\tran X)]
\label{eq:CF}.
\end{equation}

\begin{highlight}[label={yb:SuperResolution}]{Super-Resolution Recovery}
Super-resolution (SR) is a general class of techniques to enhance the resolution of a sensing system, \eg to observe sub-wavelength features in astronomy or medical imaging \cite{Candes2014}. The problem addressed by SR is to recover a continuous-time (when dimension $d=1$), or continuous-space (when dimension $d \geq 2$) sparse signal $s(\vt)$ from a few, possibly noisy Fourier measurements $\{y_j\}_{j=1}^m$.
This amounts to recovering a weighted sum $s(\vt) = \sum_{\ell=1}^k \alpha_\ell \delta(\vt-\vt_\ell)$ of $k$ Diracs with amplitudes $\alpha_\ell$ and locations $\vt_\ell\in\mathbb{R}^d$ from 
\begin{align}
y_j = \int_{\bb R^d} s(\vt) \exp(-\jmath 2\pi \vw_j\tran \vt)  \dif\vt + e_j,
\qquad j=1,\dots,m,
\label{eq:SuperResolution}
\end{align}
with measurement noise $e_j$ and frequency vectors 
$\{\vw_j\}_{j=1}^m$ in $\mathbb{R}^d$.
Recovery of $s(\cdot)$ can be posed as an infinite-dimensional convex problem on measures \cite{Candes2014} \ocite{Bredies2013}. 
However, 
most reconstruction algorithms involve non-convex steps \cite{Denoyelle2018}. 
When the frequencies $\vw_j$ are drawn randomly, the signal can be accurately recovered with high probability when $m$ is of the order of at least $kd^3$, up to log factors.
Proving this usually requires additional assumptions, such as a minimal separation between the locations $\vt_\ell$  \cite{Poon2020} or positivity of the amplitudes $\alpha_\ell$ \cite{Denoyelle2018}.

The link between super-resolution and 
\modifiedtext compressive \unmodifiedtext 
clustering follows from rewriting~\eqref{eq:RFFsketch}
as
\begin{align}
\tilde{z}_j = \int_{\bb R^d} p_X(\vx) \exp(-\jmath 2\pi \vw_j\tran \vx) \dif\vx + e_j,
\qquad j=1,\dots,m,
\label{eq:RFFsketch2}
\end{align}
where $e_j$ captures the ``noise'' due to finite-sample effects (recall~\eqref{eq:noise}).
Comparing~\eqref{eq:RFFsketch2} to~\eqref{eq:SuperResolution}, 
we see that they are mathematically equivalent
when $p_X(\vx)=\sum_{\ell=1}^k \alpha_\ell \delta(\vx-\vc_\ell)$,
except for the fact that, in the case of 
\modifiedtext compressive clustering, \unmodifiedtext
$\alpha_\ell$ are non-negative and sum to one.
\end{highlight}

Intuitively, when a probability distribution $p$ has a ``simple'' structure, one can recover it (with high probability) from 
enough randomly chosen samples of its Fourier transform. 
Centroid recovery from the sketch 
\eqref{eq:RFFsketch} is premised on the empirical distribution $\hat{p}_{\ds}$ being well approximated by a mixture of $k$ Diracs, \ie $\hat{p}_{\ds}(\vx)\approx \sum_{\ell=1}^{k} \alpha_{\ell} \delta(\xbf-\cbf_{\ell})$. 
In this case, centroid recovery 
parallels the ``super-resolution'' recovery problem (see~\ybref{yb:SuperResolution})
through an optimization problem that is the continuous analog to~\eqref{eq:KmeansBinned} (or a convex or greedy relaxation for the continuous case) and  will be further elaborated in the next section.

In both problems, recovery guarantees are possible when the frequencies $\vw_j$ are randomly drawn. For example, when the centroids $\{\cbf_\ell\}_{\ell=1}^k$ are $\epsilon$-separated and $r$-bounded, centroid recovery
guarantees have been established provided the sketch dimension $m$ is on the order of $k^{2}d \log (r/\epsilon)$, omitting---for simplicity---some log factors involving $k$ and $d$ \cite{gribonval2017}.
Similar guarantees hold when $\hat{p}_{\ds}$ is approximately a sum of spatially localized components (\eg in 
\modifiedtext compressive \unmodifiedtext 
GMM)
\cite{gribonval2017}.

\section{Learning from a sketch}\label{sec:learningfromsketch}

Until now, we have primarily focused on the first stage of the \modifiedtext compressive\unmodifiedtext-learning pipeline (see Fig.~\ref{fig:framework}), where the dataset $\ds$ is sketched down to $\empts$, a compressed and noisy representation of the underlying data-generating distribution $p_X$.
We now discuss the second stage of the pipeline, where the distributional parameters of interest, $\thetabf$, are recovered from the sketch $\empts$.
The close analogy between sketching and CS allows us to cast this ``parameter learning'' stage as an optimization problem, \ie
\begin{equation}\label{eq:learngeneric}
\widetilde \thetabf = \argmin_\thetabf C(\thetabf \smid  \empts),
\end{equation}
where the cost function $C(\cdot \smid  \empts)$ is adapted to the considered learning task.
As in CS, many candidate distributions $p_X$ 
 (and hence many candidate parameters $\thetabf$)
 can yield the same sketch $\empts$. 
Thus, to make the inverse problem well-posed, one needs to employ \emph{concrete modeling assumptions} and \emph{regularization}, both of which can take several forms. 
As in CS, we will assume that the sketched quantity $p_X$ is of low intrinsic complexity, \ie close to some family of ``simple'' probability distributions.  

As a first example, we consider the problem of learning a mixture model from a sketch. 
Similar to a sparse vector $\Vec{x}$, which is a linear combination of a few elements of the standard basis,  
a \emph{mixture model} $p_X$ is a linear combination of a few ``simple'' densities $\{p_{\thetabf_\ell}\}_{\ell=1}^k$.
Concretely, $p_X=\sum_{\ell=1}^k \alpha_\ell p_{\thetabf_\ell}$, where the mixture weights $\{\alpha_\ell\}_{\ell=1}^k$ are non-negative and sum to one.
For example, with a Gaussian Mixture Model (GMM), we have that $p_{\thetabf_\ell} = \cN(\mubf_\ell, \Sigmabf_\ell)$, where $\thetabf_\ell = \{\mubf_\ell, \Sigmabf_\ell\}$ contains the mean $\mubf_\ell$ and covariance $\Sigmabf_\ell$. 
If $p_X$ is well approximated by the mixture model $\sum_{\ell=1}^k \alpha_\ell p_{\thetabf_\ell}$, then, according to~\eqref{eq:sketch-linear-projection}, the sketch $\empts$ is well approximated by the linear combination $\sum_{\ell=1}^{k} \alpha_{\ell} \cA(p_{\thetabf_{\ell}})$.
Hence, one could try to extract the mixture parameters, $\thetabf=\{\alpha_{\ell},\thetabf_\ell\}_{\ell=1}^k$, from the sketch $\empts$ by solving the (nonconvex) optimization problem~\eqref{eq:learngeneric} with
\begin{align}
C(\thetabf\smid \empts) := 
\bigg\| \empts - \sum_{\ell=1}^k \alpha_\ell \cA(p_{\thetabf_\ell}) \bigg\|^2
\label{eq:learnmixture}.
\end{align}
The cost $C(\thetabf \smid  \empts)$ can be interpreted as the negative log-likelihood (up to a shift and scale) of $\thetabf$ given the sketch $\empts$, under the classic modeling assumption of \iid Gaussian measurement noise $\noise$ in~\eqref{eq:sketch-linear-projection}.

When the RF feature map $\fmap(\cdot)$ is used to compute the sketch $\empts$ and the component densities $p_{\thetabf_{\ell}}$ are Gaussian or $\alpha$-stable, there exist analytic expressions for $\cA(p_{\thetabf_{\ell}})$ and for the gradient of $\cA(p_{\thetabf_{\ell}})$ with respect to the mixture parameters in $\thetabf_{\ell}$ \cite{keriven2017b}.
These expressions are convenient when numerically optimizing~\eqref{eq:learnmixture}.
For instance, greedy approaches, similar to the OMP algorithm for CS (recall~\ybref{yb:CS}), can be used \cite{keriven2017b} to estimate the parameters $\{\alpha_\ell,\thetabf_\ell\}_{\ell=1}^k$.  
These approaches sequentially estimate and subtract, from $\empts$, each of the $k$ components $\alpha_\ell \cA(p_{\thetabf_{\ell}})$ that best align with it, where ``best'' is measured via the correlation between $\empts$ and $\cA(p_{\thetabf_{\ell}})$.
As another example, an iterative approach \cite{byrne2019sketched} was proposed that exploits the log-likelihood interpretation of $C(\thetabf\smid \empts)$ in~\eqref{eq:learnmixture} and the \iid random nature of the linear transform $\linop$ in the RF map~\eqref{eq:RFF}.
\ifarxiv
In \cite{keriven2017b} and \ocite{Keriven2018b},
\fi
applications of 
\modifiedtext compressive mixture-modeling
\unmodifiedtext
are demonstrated on speaker verification (see~\ybref{yb:audio}) and source separation.

As a second example, we consider 
\modifiedtext compressive \unmodifiedtext
\kmeans clustering. 
Here, the goal is to recover, from the sketch $\empts$, the centroids $\thetabf=\{\cbf_\ell\}_{\ell=1}^k$ that minimize the average squared Euclidean distance from each sample to its nearest centroid, \ie $\tfrac{1}{n}\sum_{i=1}^n \min_{\ell} \|\xbf_i-\cbf_\ell\|^2$.
To tackle this \kmeans problem, we view it as an approximation of a particular GMM fitting problem.
In particular, suppose that the probability distribution $p_X$ is a GMM with weights $\alpha_\ell$, mean vectors $\cbf_\ell$, and covariance matrices $\Sigmabf_\ell$. 
\modifiedtext
Then, in the special case that $\alpha_\ell=1/k$ and $\Sigmabf_\ell=\sigma^2\Ibf$ for all components $1\leq \ell \leq k$, 
we can write the likelihood as
$\prod_{i=1}^n p(\xbf_i|\thetabf) 
\propto \prod_{i=1}^n \sum_{\ell=1}^{k} \exp(-\frac{1}{2\sigma^2}\|\xbf_i-\cbf_l\|^2)$,
and so the negative log-likelihood becomes 
$-\sum_{i=1}^{n} \log \sum_{\ell=1}^{k} \exp(-\frac{1}{2\sigma^2}\|\xbf_i-\cbf_l\|^2)$ 
up to an additive constant.
We can then use the log-sum-exp approximation 
$\log \sum_\ell \exp(f_\ell(\xbf)) \approx \max_\ell f_\ell(\xbf)$
to approximate this latter expression as
\unmodifiedtext
$\frac{1}{2\sigma^2}\sum_{i=1}^n \min_\ell \|\xbf_i-\cbf_\ell\|^2$,
which agrees with the \kmeans cost up to a scaling.
If we furthermore consider the case of a vanishing variance $\sigma^2\rightarrow 0$, then the component density $p_{\thetabf_\ell}$ reduces to a point mass, \ie $p_{\thetabf_\ell}(\xbf) \rightarrow \delta(\xbf-\cbf_\ell)$.
In this limiting case, the linear measurement of the point-mass $p_{\thetabf_\ell}$ is
$\cA(p_{\thetabf_\ell})
 =\Exp_{X\sim p_{\thetabf_\ell}}[\fmap(X)]
 =\int \fmap(\xbf) p_{\thetabf_\ell}(\xbf) \dif\xbf 
 =\fmap(\cbf_\ell)$
according to~\eqref{eq:ProjDef}.

Thus, with these justifications, the cost function~\eqref{eq:learnmixture} for 
\modifiedtext compressive \unmodifiedtext
GMM would change to
\begin{align}
C(\thetabf \smid  \empts) := 
 \bigg\| \empts - \frac{1}{k}\sum_{\ell=1}^k \fmap(\cbf_\ell) \bigg\|^2
\label{eq:learnkmeansextended} 
\end{align}
for 
\modifiedtext compressive \unmodifiedtext
\kmeans.
If we do not want to assume that $\alpha_\ell=1/k$ for each $\ell$, we could instead estimate $\{\alpha_\ell\}_{\ell=1}^k$ from the sketch, leading to the cost function  
suggested in \cite{keriven2017a}:
\begin{align}
C(\thetabf \smid  \empts) := 
\min_{\boldsymbol{\alpha}} \bigg\| \empts - \sum_{\ell=1}^k \alpha_\ell \fmap(\cbf_\ell) \bigg\|^2
\label{eq:learnkmeans}.
\end{align}
Similar to the 
\modifiedtext compressive \unmodifiedtext
GMM problem described earlier, 
minimization of the \modifiedtext cost function~\eqref{eq:learnkmeans} for compressive \kmeans \unmodifiedtext can be tackled by greedy approaches, as described in \cite{keriven2017a}.
Despite the fact that~\eqref{eq:learnkmeans} does not directly minimize the \kmeans cost, it has been shown empirically \cite{keriven2017a} that the centroids estimated by such greedy algorithms nearly minimize this cost, see, \eg~\ybref{yb:MNIST} for an example on MNIST data. 
This claim is also supported by theoretical results guaranteeing that the minimizer of~\eqref{eq:learnkmeans} is endowed with statistical-learning guarantees with respect to the original \kmeans cost~\cite{gribonval2017}. 
Such guarantees will be discussed shortly.

Depending on the choice of parameters $\thetabf$ and the feature map $\fmap(\cdot)$, 
\modifiedtext the form of the optimization problem posed to learn-from-a-sketch
\unmodifiedtext
can differ considerably from that for GMMs in~\eqref{eq:learnmixture} and that for \kmeans in~\eqref{eq:learnkmeansextended}-\eqref{eq:learnkmeans}.
Consider, for example, 
\modifiedtext learning-from-a-sketch for PCA.
\unmodifiedtext
As described earlier, the parameter $\thetabf$ of interest is the $k$-dimensional subspace that best fits the $d$-dimensional data $\{\vx_i\}_{i=1}^n$ in a least-squares sense.
It is well known that this subspace is spanned by $k$ principal eigenvectors of the empirical autocorrelation matrix $\hat{\Rbf} = \frac{1}{n}\sum_{i=1}^n \xbf_i \xbf_i\tran$, or---equivalently---the column space of the (symmetric \finalrevisions{positive semi-definite}) matrix $\hat{\Rbf}_k$ that is closest to $\hat{\Rbf}$ in the Frobenius norm: 
\begin{equation}
\hat{\Rbf}_k
:=\argmin_{\Rbf:\,\rank(\Rbf)\leq k} \|\hat{\Rbf}-\Rbf\|_F 
=\argmin_{\Rbf:\,\rank(\Rbf)\leq k} \|\vect(\hat{\Rbf})-\vect(\Rbf)\|^{2} 
\label{eq:learnPCA0} .
\end{equation}
When sketching using quadratic features 
$\fmap(\vx) = ( (\wbf_{1}\tran\vx)^{2},\ldots, (\wbf_{m}\tran\vx)^{2})$
with random $\wbf_j \in \mathbb{R}^d$, the $j$th component of the sketch becomes $\tilde{z}_j = \frac{1}{n} \sum_{i=1}^n \wbf_j\tran \vx_i\vx_i\tran \wbf_j = \wbf_j\tran\hat{\Rbf} \wbf_j$. 
Importantly, this $\tilde{z}_j$ is a linear function of $\hat{\Rbf}$, and so there exists an $m \times d^2$ matrix $\Abf$ such that $\empts = \Abf \vect(\hat{\Rbf})$. 
Thus, by analogy with~\eqref{eq:learnPCA0}, one could first fit a \finalrevisions{symmetric positive semi-definite} low-rank matrix to the sketch $\empts$ via
\begin{equation}
\widetilde{\Rbf} = \argmin_{\Rbf:\,\rank(\Rbf) \leq k,\, \Rbf\tran=\Rbf, \finalrevisions{\Rbf \succeq 0}}  \|\empts-\Abf \vect(\Rbf)\|^{2}
\label{eq:learnPCA} ,
\end{equation}
and then set the parameter estimate $\widetilde{\thetabf}$ equal to the column space of $\widetilde{\Rbf}$. \finalrevisions{While the optimum of~\eqref{eq:learnPCA0} is automatically symmetric and positive semi-definite, this property needs to be enforced explicitly in~\eqref{eq:learnPCA}.}

The low-rank matrix recovery problem~\eqref{eq:learnPCA} has been thoroughly investigated by the signal processing and machine learning communities (\eg \cite{chenchi2018}). 
It arises, for example, in applications such as collaborative filtering for recommender systems and signal reconstruction from phaseless measurements. 
Early approaches to solving the non-convex problem~\eqref{eq:learnPCA} involved convex relaxation via nuclear-norm regularization \cite[Chapter 4]{FouRau13}.
More recent approaches exploit the non-convex geometry of~\eqref{eq:learnPCA} \cite{NIPS2016_6048}.


\section{
\modifiedtext Compressive \unmodifiedtext 
learning with theoretical guarantees} 

In the previous section, 
\modifiedtext learning-from-a-sketch was posed
\unmodifiedtext
as the optimization problem~\eqref{eq:learngeneric}, repeated here for convenience:
\begin{equation}\label{eq:learngeneric2}
 \widetilde \thetabf = \argmin_\thetabf C(\thetabf \smid  \empts) .
\end{equation}
The quantity $C(\cdot\smid  \empts)$ is a real-valued \emph{cost function} whose minimizer $\widetilde{\thetabf}$ is the ``best'' (in some sense) estimate of $\thetabf$ from the sketch $\empts$. 
This approach contrasts with traditional statistical learning~\cite{vapnik2013nature,ShalevShwartz:2009jm}, which computes the parameter estimate
\begin{equation}\label{eq:learntraditional}
\hat{\thetabf} := \argmin_{\thetabf} R(\thetabf\smid \ds)
\qquad \text{with}\
R(\thetabf \smid  \ds) :=\frac{1}{n}\sum_{i=1}^n L(\thetabf\smid \xbf_i),
\end{equation}
where $R(\cdot\smid \ds)$ is the ``empirical risk'' and $L(\cdot\smid \xbf_i)$ is the ``loss function'' for the $i$th sample. 
Table~\ref{tab:loss-function-four-working-examples} presents typical loss functions for our four running examples.
In most cases, performing the minimization in~\eqref{eq:learntraditional} involves querying the full training dataset $\ds$ many times, \eg for stochastic gradient descent. 
In 
\modifiedtext compressive \unmodifiedtext 
learning, the cost $C(\cdot\smid \empts)$, which depends only on the low-dimensional sketch $\empts$, is used as a \emph{surrogate} for the empirical risk.
Because the dimension of the sketch is so much smaller than the cardinality of the full dataset $\ds$, the minimization in~\eqref{eq:learngeneric} can be made much more efficient than that in~\eqref{eq:learntraditional}.

\begin{table}[!t]
  \centering
  \caption{Loss functions for our four running examples}
  \label{tab:loss-function-four-working-examples}
  \scalebox{0.85}{
    \begin{tabularx}{1.1\linewidth}{lll}
    \toprule
    Running example&Parameters $\thetabf$&Loss function $L(\thetabf\smid  \vx_i)$\\
    \midrule
    Principal component analysis&Orthonormal basis $\thetabf = \{\ubf_\ell\}_{\ell=1}^k$&$\sum_{\ell=1}^k |\vx_i\tran \ubf_\ell|^2$\\
    Least-squares linear regression&A weight matrix $\regmat$&$\|\vx_{1i} - \regmat \vx_{2i}\|^2$, $\vx_i := (\vx_{1i}\tran,\vx_{2i}\tran)\tran$\\
    Gaussian mixture modeling&GMM parameters $\thetabf = \{\alpha_\ell,\bs\mu_\ell,\bs\Sigma_\ell\}_{\ell=1}^k$&\parbox{5cm}{$-\log \sum_{\ell=1}^k \alpha_\ell \cl N(\vx_i;\bs \mu_\ell,\bs \Sigma_\ell)$}\\
    \kmeans clustering&A set of centroids $\thetabf = \{\bs c_\ell\}_{\ell=1}^k$&$\min_{\ell} \|\vx_i - \bs c_\ell\|^{2}$\\
    \bottomrule
    \end{tabularx}
  }
\end{table}

Of course, the estimate \modifiedtext $\widetilde \thetabf$, which \unmodifiedtext minimizes the cost $C(\cdot\smid \empts)$ is not, in general, the same as \modifiedtext $\widehat \thetabf$, which \unmodifiedtext minimizes the empirical risk $R(\cdot\smid \ds)$.
Both, however, are approximations of the \emph{ideal} estimate
\begin{equation}\label{eq:learnideal}
\thetabf^\star := \argmin_{\thetabf} R^\star(\thetabf) 
\qquad \text{with}\
R^\star(\thetabf) := \bE [L(\thetabf\smid  X)],
\end{equation}
where $R^\star(\thetabf)$ is known as the ``true risk.'' 
Indeed, the feature map $\fmap(\cdot)$ and cost $C(\cdot \smid \empts)$ are designed precisely to ensure that 
\modifiedtext the surrogate \unmodifiedtext minimization~\eqref{eq:learngeneric2} approximates the true-risk minimization~\eqref{eq:learnideal}. 

\subsection{\modifiedtext Excess risk guarantees \unmodifiedtext}    
\modifiedtext To establish a guarantee on the goodness of an estimate $\thetabf$, one must prove that the ``excess risk'' $R^\star(\thetabf)-R^\star(\thetabf^\star) \geq 0$ is small. Indeed, the \unmodifiedtext
true risk can be interpreted as the \emph{expected} loss on test samples that have the same generating distribution $p_X$ as the training samples $\ds$, but that are drawn independently and not accessible at training time. 
In statistical learning tasks, the true risk $R^\star(\cdot)$ is the primary metric by which one judges the quality of an arbitrary estimate $\thetabf$.
Note that 
\modifiedtext controlling the excess risk \unmodifiedtext
is different from proving that $\thetabf$ is close to the ideal estimate $\thetabf^\star$ in, \eg the Euclidean distance $\|\thetabf-\thetabf^\star\|$. 
Consider, for example, the problem of fitting a one-dimensional linear subspace to a dataset (\ie PCA with $k=1$).  
When the two largest eigenvalues of the autocorrelation matrix $\Rbf$ are equal, any nontrivial linear combination of the corresponding eigenvectors generates a one-dimensional subspace $\thetabf$ with minimum true risk. 
The problem of estimating a subspace ``close to the optimal one'' is thus ill-defined, yet finding a subspace with close-to-optimal performance is well-defined and achievable, both by classic PCA and 
\modifiedtext compressive \unmodifiedtext 
PCA.

\begin{highlight}[label={yb:LvsR}]{Traditional Statistical Learning versus Compressive Learning}

{
\begin{center} 
  \includegraphics[width=0.95\textwidth]{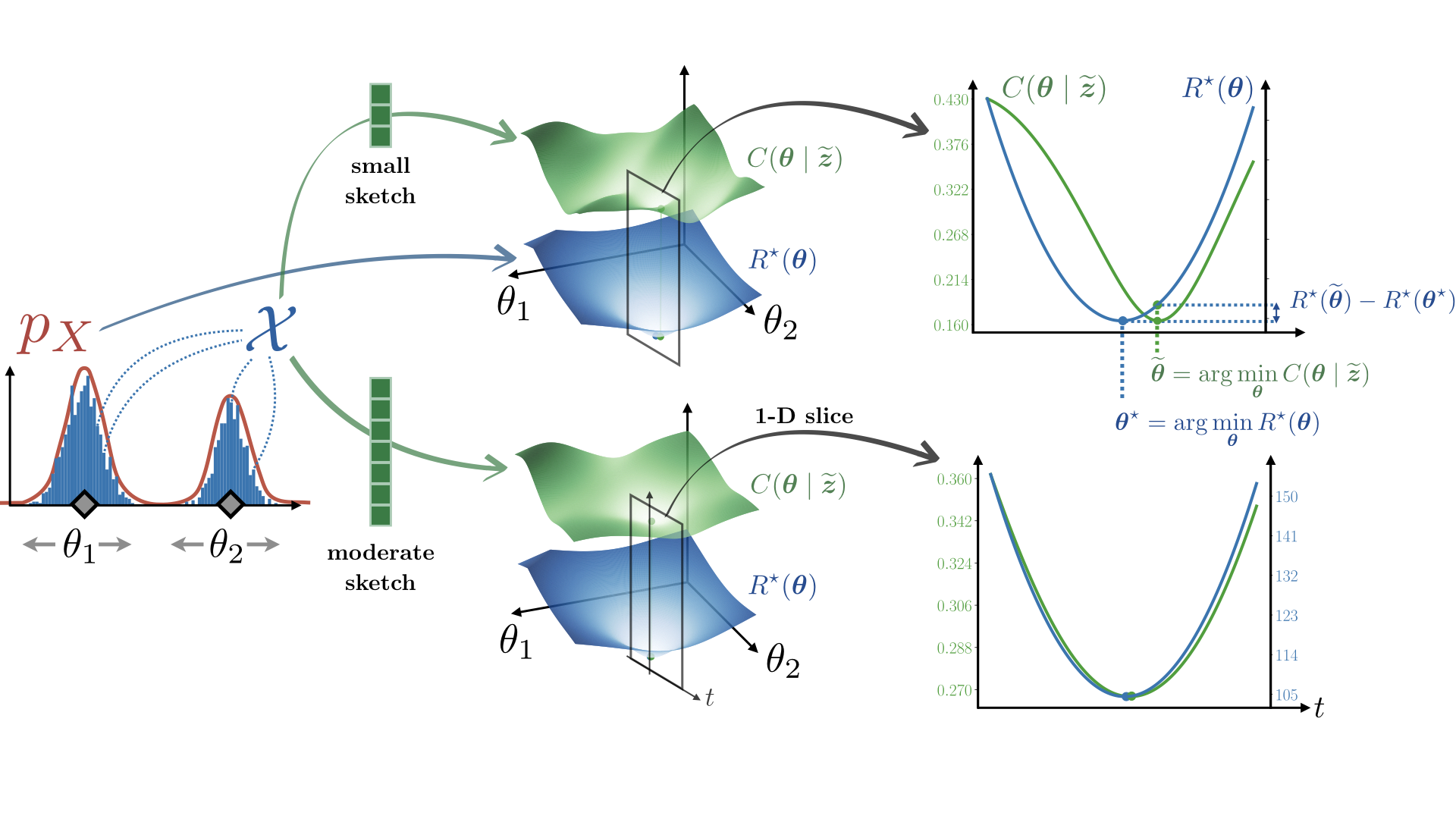}
  \parbox{0.8\textwidth}{\captionof{figure}{\small \kmeans clustering of one-dimensional data $\ds$, showing cost function $C(\thetabf\smid \empts)$, risk $R^\star(\thetabf)$, ideal estimate $\thetabf^\star$, \modifiedtext compressive \unmodifiedtext 
learning estimate $\tilde{\thetabf}$, and excess risk $R^\star(\tilde{\thetabf})-R^\star(\thetabf^\star)$, for a sketch of small dimension $m$ (top) and \modifiedtext for a sketch of moderate size \unmodifiedtext $m$ (bottom).}\label{fig:LvsR}}
\end{center}
}

In statistical learning, the ideal parameter  
$\thetabf^{\star}$ minimizes the true risk $R^{\star}( \thetabf) = \bE_{X\sim p_X} [L(\thetabf\smid  X)]$.
The performance of any estimate $\thetabf$ is measured according to its excess risk, \ie $R^{\star}(\thetabf)-R^{\star}(\thetabf^{\star})$. 
In 
\modifiedtext compressive \unmodifiedtext
learning, one obtains $\tilde{\thetabf}$ by minimizing a cost function $C(\thetabf\smid \empts)$, where the sketch $\empts$ is a compressed version of a finite-size dataset $\ds$ with samples drawn from distribution $p_X$.

To gain intuition into how these quantities manifest in 
\modifiedtext compressive \unmodifiedtext
learning, Fig.~\ref{fig:LvsR} shows a simple example:
\modifiedtext compressive \unmodifiedtext
\kmeans clustering of one-dimensional data $\ds=\{x_i\}_{i=1}^n$ with two centroids, $\theta_1$ and~$\theta_2$. 
The true risk $R^{\star}(\cdot)$ and the cost $C(\cdot\smid \empts)$ are plotted versus $\thetabf = (\theta_{1},\theta_{2})$ and versus a 1-D slice in the $\thetabf$ plane.
As the sketch dimension $m$ increases, it can be seen that 
the excess risk $R^{\star}(\widetilde{\thetabf}) - R^{\star}(\thetabf^{\star})$ decreases.
Moreover, although the cost function $C(\cdot\smid \empts)$ is nonconvex, 
it is approximately quadratic in a large basin of attraction around its global minimizer, suggesting that gradient-descent algorithms will behave well when properly initialized.

\end{highlight}

Despite the fact that the estimates $\widetilde{\thetabf}$, $\hat{\thetabf}$, and $\thetabf^{\star}$ 
minimize different objective functions, they often yield similar true-risks, \ie the excess risks of $\widetilde{\thetabf}$ and $\hat{\thetabf}$ are provably small. 
For $\hat{\thetabf}$, the proofs use classic results from statistical learning \cite{vapnik2013nature,ShalevShwartz:2009jm}, under assumptions that we will briefly discuss in the sequel. 
For $\widetilde{\thetabf}$, with an appropriately designed feature map $\fmap(\cdot)$ and cost function $C(\cdot \smid \empts)$, the proofs informally follow from the fact that $C(\cdot\smid  \empts)$ and $R^{\star}(\cdot)$ have a similar shape. 
This fact is illustrated in~\ybref{yb:LvsR} for a toy example of 
\modifiedtext compressive \unmodifiedtext 
clustering. 
There it can be seen that, with 
a properly designed feature map $\fmap(\cdot)$, 
a well-chosen cost function $C(\cdot\smid \empts)$, and
a sufficiently large sketch dimension $m$, 
the minimizer $\tilde{\thetabf}$ of the cost yields a nearly minimal true risk $R^{\star}(\tilde{\thetabf})$ and lives in a large basin-of-attraction in $C(\cdot\smid \empts)$. 
If the sketch dimension $m$ is chosen too small, however, then the excess risk $R^{\star}(\tilde{\thetabf})-R^{\star}(\thetabf^{\star})$ increases.

A theory of 
\modifiedtext compressive \unmodifiedtext 
learning~\cite{gribonval2017} has been developed to better understand how (random) feature maps $\fmap(\cdot)$ and cost functions $C(\cdot\smid \empts)$ can be designed to ensure that the sketch $\empts$ captures sufficient information 
\modifiedtext 
to control the excess risk on $\tilde{\thetabf}$.
\textbf{In the following four subsections, we aim to summarize the key aspects of this theory using broadly accessible language.}
For those interested in the full technical details, they can be found in \cite{gribonval2017}.

In a nutshell, the theory relies on interpreting $\tilde{\thetabf}$ 
as the minimizer of the risk 
attributed to a \emph{surrogate} probability distribution $\tilde{p}$ 
with the following properties:
\begin{itemize}
    \item $\tilde{p}$ is ``close'' to the distribution $p_X$, as measured by a task-driven distance $d(\tilde{p},p_X)$;
    \item $\tilde{p}$ is a ``simple'' distribution (\eg for compressive GMM, $\tilde{p}$ is a Gaussian mixture);
    \item $\tilde{p}$ minimizes $\|\empts - \skop(p)\|$ among all simple distributions $p$.
\end{itemize}
For example, in compressive GMM, this holds when $\tilde{p} = p_{\tilde{\thetabf}}$, \ie the Gaussian mixture distribution parameterized by $\tilde{\thetabf}$.
Given such a $\tilde{p}$, the theory can be explained in the following step-by-step manner:
\begin{itemize}
\item
Just as in traditional statistical learning, excess risk bounds can be established if the task-driven distance $d(\tilde{p},p_X)$ is controlled (see Section~\ref{sec:distance}). 
\item
Given a family of ``simple'' probability distributions (see Section~\ref{sec:modelset}), this distance can indeed be controlled provided a certain ``lower restricted isometry property'' (LRIP) holds (see Section~\ref{sec:LRIPsufficient}). 
This is analogous to traditional CS, where the RIP allows one to control the performance of sparse signal recovery.
\item
To establish the LRIP for a random feature maps $\fmap(\cdot)$, one can build on connections between kernel methods and the JL Lemma (see Section~\ref{sec:kernelJL}). 
\item
Finally, to design task-specific feature maps $\fmap(\cdot)$, one can appeal to the problem of kernel design (see Section~\ref{sec:design}).
\end{itemize}
\unmodifiedtext

\subsection{Task-driven distances and excess-risk bounds}
\label{sec:distance}
In traditional statistical learning, it is common to define a distance between the true distribution $p_X$ and an arbitrary surrogate $p'_X$ as
\begin{equation}
d(p'_{X},p_{X}) = \sup_{\thetabf}\big|R^{\star}(\thetabf\smid  p'_{X})-R^{\star}(\thetabf\smid  p_{X})\big|
\label{eq:distance},
\end{equation}
where $R^{\star}(\thetabf\smid  p) := \bE_{X \sim p} [L(\thetabf\smid  X)]$ denotes the risk under an arbitrary distribution $p$.
This distance is task specific, since the loss function $L(\cdot\smid \cdot)$ depends on the estimation task.
The utility of $d(p'_{X},p_{X})$ for assessing the goodness of parameter estimation follows from the inequalities
\begin{align}
R^{\star}(\thetabf^{\star}\smid  p_{X}) 
\leq R^{\star}(\thetabf^{\star\prime}\smid  p_{X}) &\leq R^{\star}(\thetabf^{\star\prime}\smid  p'_{X}) + d(p'_{X},p_{X}) \notag\\
&\leq R^{\star}(\thetabf^{\star} \smid  p'_{X}) + d(p'_{X},p_{X}) \notag\\
&\leq R^{\star}(\thetabf^{\star} \smid  p_{X})+ 2 d(p'_{X},p_{X}) , \label{eq:riskinequalities}
\end{align}
where $\thetabf^\star:=\argmin_{\thetabf}R^\star(\thetabf|p_X)$
and $\thetabf^{\star\prime}:=\argmin_{\thetabf}R^\star(\thetabf|p_X')$.
In particular, these inequalities yield the following bounds on the excess risk of $\thetabf^{\star\prime}$:
\begin{equation}
0 \leq R^{\star}(\thetabf^{\star\prime} \smid  p_{X}) -R^{\star}(\thetabf^{\star} \smid  p_{X}) \leq 2 d(p'_{X},p_{X})
\label{eq:excessriskbound}.
\end{equation}

For example, if $p'_X$ equals the empirical distribution $\hat{p}_{\ds}(\xbf) := \frac{1}{n}\sum_{i=1}^n \delta(\xbf-\xbf_i)$, then~\eqref{eq:excessriskbound} bounds the ``generalization error'' of \emph{empirical risk minimization}, \ie the error incurred when training with $\ds=\{\xbf_i\}_{i=1}^n$ but testing with independent samples drawn from $p_X$. 
This is the reasoning behind many classic statistical-learning guarantees, where so-called ``uniform convergence'' results establish that, under appropriate conditions, $d(\hat{p}_{\ds},p_{X}) =O(1/\sqrt{n})$ 
with high probability on the draw of \iid training samples $\ds=\{\xbf_i\}_{i=1}^n$. 
The term ``uniform convergence'' stems from the analogy between the distance~\eqref{eq:distance} and the $\ell_{\infty}$ norm.

\modifiedtext 
\emph{Similarly, excess risk bounds for compressive learning are achieved by interpreting the minimizer $\tilde{\thetabf}$ of \eqref{eq:learngeneric2} as the minimizer of the risk $R^\star(\thetabf|\tilde{p})$, for some ``simple'' probability distribution $\tilde{p}$ that minimizes $\|\empts-\skop(\tilde{p})\|$, and by controlling the distance $d(\tilde{p},p_X)$. 
To control this distance, key geometric intuitions exploit the connections between compressive learning and CS, as illustrated in Fig.~\ref{fig:skvscs}.  } 
\unmodifiedtext
\subsection{Exploiting simplified models to learn from a sketch}
\label{sec:modelset}


Recall that, in CS, the goal is to recover the best $k$-sparse approximation of an unknown vector $\xbf_0\in\mathbb{R}^d$ given the noisy linear measurement $\ybf=\Abf\xbf_0+\noise\in\mathbb{R}^m$, where $m\ll d$.
Ideally, recovery aims to find the $k$-sparse vector whose (noiseless) measurement is closest to the observed $\ybf$, \ie \begin{equation}\label{eq:CS}
\hat{\xbf} = \argmin_{\xbf\in\Sigma_k} \|\ybf-\Abf\xbf\|^{2} .
\end{equation}
In practice, convex or greedy approximations of this 
combinatorial approach are often used.
Recovery guarantees can be established using the \emph{restricted isometry property} (RIP) in~\eqref{eq:RIP}, which says that the Euclidean distance $\|\Abf \xbf-\Abf \xbf'\|$ between the (noiseless) measurements of two $k$-sparse vectors is almost the same as the Euclidean distance $\|\xbf-\xbf'\|$ between the vectors themselves (see~\ybref{yb:CS}).
These guarantees require that 
the sparsity $k$ is sufficiently small for a given $m$ and $d$.

To connect 
\modifiedtext compressive \unmodifiedtext
learning to CS, we first reframe 
\modifiedtext the goal of learning-from-a-sketch as that of \unmodifiedtext
 recovering a parametric model \emph{distribution} $p_{\thetabf}\in\Sigma_{\Thetabf}$ from the sketch $\empts$, rather than recovering the model parameters $\thetabf\in\Thetabf$ themselves.
Here, $\Sigma_{\Thetabf}:=\{p_{\thetabf}: \thetabf\in\Thetabf\}$ denotes some set of admissible distributions $p_{\thetabf}$.
For example, in 
\modifiedtext compressive \unmodifiedtext
GMM, the goal becomes that of recovering a $k$-term GMM $p_\thetabf=\sum_{\ell=1}^k\alpha_\ell \cN(\mubf_\ell, \Sigmabf_\ell)$ rather than the GMM parameters $\thetabf=\{\alpha_\ell,\mubf_\ell,\Sigmabf_\ell\}_{\ell=1}^k$.
Likewise, in 
\modifiedtext compressive \unmodifiedtext
PCA, the goal becomes that of recovering a distribution $p_\thetabf$ whose autocorrelation matrix has rank of at most $k$, rather than a $k$-dimensional orthonormal basis $\thetabf$ for the column space of that autocorrelation matrix.
Then, mirroring the CS recovery approach \eqref{eq:CS}, \modifiedtext compressive \unmodifiedtext
learning recovery aims to find a parametric distribution $p_{\thetabf}\in\Sigma_\Thetabf$ whose (noiseless) sketch is closest to the observed sketch $\empts$, \ie 
\begin{equation}
\modifiedtext \tilde{p} = \unmodifiedtext 
p_{\tilde{\thetabf}} 
:= \argmin_{p_{\thetabf}\in\Sigma_\Thetabf} \|\empts-\cA(p_{\thetabf})\|^{2}
\label{eq:SketchFitting}.
\end{equation}
The construction of $\Sigma_\Thetabf$ implies that the parameters $\tilde{\thetabf}$ defining $p_{\tilde{\thetabf}}$ minimize a certain cost function, \ie
\begin{equation}
\tilde{\thetabf}
= \argmin_{\thetabf\in\Thetabf} C(\thetabf \smid  \empts) 
\qquad \text{with}\
C(\thetabf \smid  \empts)=\|\empts-\cA(p_{\thetabf})\|^{2} 
\label{eq:SketchFitting2}.
\end{equation}
Note the similarity between~\eqref{eq:SketchFitting2} and~\eqref{eq:learngeneric}-\eqref{eq:learnmixture}.
\modifiedtext 
Moreover, $\tilde{\thetabf}$ also minimizes the risk $R^\star(\thetabf\smid\tilde{p})$ \cite{gribonval2017}.
\unmodifiedtext

From the description above, we see that a central theme of both CS and 
\modifiedtext compressive \unmodifiedtext
learning is that \emph{measurement compression makes it impossible to recover the full object-of-interest} (\ie $\xbf_0$ in CS or $p_X$ in 
\modifiedtext compressive \unmodifiedtext
learning) without additional side-information. 
For this reason, both seek to recover the parameters of \emph{a simplified model} of the object-of-interest. 
In CS, this is accomplished by seeking to recover the best $k$-sparse approximation to $\xbf_0$, rather than $\xbf_0$ itself.
In 
\modifiedtext compressive \unmodifiedtext
learning, this is accomplished by seeking to recover the best model distribution $p_{\thetabf}\in\Sigma_{\Thetabf}$, rather than the true distribution $p_X$.
In both cases, if the linear operator (\ie $\Abf$ in CS or $\cA(\cdot)$ in 
\modifiedtext compressive \unmodifiedtext 
learning) is well designed, then 
the model parameters 
(in $\Sigma_k$ for CS or in $\Sigma_{\Thetabf}$ for 
\modifiedtext compressive \unmodifiedtext 
learning) can be accurately recovered. 

\subsection{\modifiedtext The LRIP and excess-risk control \unmodifiedtext}
\label{sec:LRIPsufficient}
Recovery guarantees can be established using a tool analogous to the RIP~\eqref{eq:RIP} in CS.

%
Take, for example, the case of 
\modifiedtext compressive \unmodifiedtext
PCA.
As discussed around~\eqref{eq:learnPCA}, one could use a sketch of the form $\empts = \Abf\vect(\hat{\Rbf})\in\mathbb{R}^m$, where $\hat{\Rbf}=\frac{1}{n}\sum_{i=1}^n \xbf_i\xbf_i\tran$ is the $d\times d$ empirical autocorrelation matrix and $m\ll d^2$, and then search for $\tilde{\Rbf}$, the symmetric matrix with rank at-most-$k$ that minimizes the cost $C(\Rbf\smid \empts)=\|\empts-\Abf\vect(\Rbf)\|^{2}$.
Recovery guarantees can be established using an RIP of the form
\begin{equation}\label{eq:lowrankRIP}
(1-\delta) \|\Rbf-\Rbf'\|_{F}^{2} 
\leq \|\Abf\vect(\Rbf)-\Abf\vect(\Rbf')\|^{2}
\leq (1+\delta) \|\Rbf-\Rbf'\|_{F}^{2},
\quad \forall \Rbf,\Rbf'\in\Sigma_k,
\end{equation}
where $\delta\in(0,1)$ is a tolerance and $\Sigma_k$ is now the set of $d\times d$ symmetric rank-$k$ matrices.
It is known \cite[Chapter 9]{FouRau13} that \iid 
\new{Gaussian} $\Abf$ satisfy the RIP~\eqref{eq:lowrankRIP} with high probability when $m$ is on the order of $kd$ times a constant that depends on the tolerance $\delta$.
Furthermore, the cost-minimizing $\tilde{\Rbf}$ is near-optimal in the sense of minimizing the excess risk, even when a convex relaxation of the cost is used \cite{gribonval2017}.

These 
\modifiedtext compressive-\unmodifiedtext
PCA guarantees hold even under \emph{model mismatch}: although $\Sigma_k$ constrains $\tilde{\Rbf}$ to have rank at-most-$k$, the empirical autocorrelation matrix $\hat{\Rbf}$ used to construct $\empts$ tends to have full rank $d$ in practice.
To explore this idea in more detail, suppose for the moment that the data $\ds$ lies in a rank-$k$ subspace. 
In this case, $\hat{\Rbf}$ would have rank at-most-$k$, and there exists a symmetric $\Rbf$ with rank at-most-$k$ that drives the cost $C(\Rbf\smid \empts)=\|\Abf\vect(\hat{\Rbf})-\Abf\vect(\Rbf)\|^{2}$ to zero.
Furthermore, when $\Abf$ satisfies the RIP~\eqref{eq:lowrankRIP}, the left inequality in~\eqref{eq:lowrankRIP} implies that this cost-minimizing $\tilde{\Rbf}$ must equal $\hat{\Rbf}$.
When the data $\ds$ does not lie in a rank-$k$ subspace, the minimal cost will be non-zero. 
In this case, an upper bound on the error $\|\hat{\Rbf}-\tilde{\Rbf}\|_F$ of the cost-minimizing estimate $\tilde{\Rbf}$, as well as an upper bound on the excess risk of the principal subspace $\tilde{\thetabf}$ of $\tilde{\Rbf}$, can both be obtained as a consequence of the RIP~\eqref{eq:lowrankRIP}.

\modifiedtext In place of the RIP, compressive \unmodifiedtext 
learning uses the so-called ``\emph{lower} RIP'' (LRIP),
\begin{equation}
d(p_{\thetabf},p_{\thetabf'}) \leq C_{0}
\|\cA(p_{\thetabf})-\cA(p_{\thetabf'})\| 
\label{eq:LRIP} ,
\end{equation}
assumed to be valid for each pair $p_{\thetabf},p_{\thetabf'} \in \Sigma_{\Thetabf}$,
where $C_0$ is a positive constant.
The LRIP says that 
the Euclidean distance $\|\cA(p_{\thetabf})-\cA(p_{\thetabf'})\|$ between the (noiseless) sketches of two distributions is---up to a scaling---controlling  the distance $d(p_{\thetabf},p_{\thetabf'})$ between the distributions themselves~\eqref{eq:distance}.
Note that the lower bound $d(\cdot,\cdot)$ in~\eqref{eq:LRIP} is task-specific, and not Euclidean as in the CS case~\eqref{eq:RIP}.

\modifiedtext The first main theoretical guarantee about compressive learning \cite{gribonval2017} is that, \unmodifiedtext
when the LRIP \eqref{eq:LRIP} holds, the estimate $\tilde{\thetabf}$ obtained by minimizing \eqref{eq:SketchFitting2} automatically
\modifiedtext has controlled excess risk. \unmodifiedtext
In particular, \modifiedtext using $\tilde{\thetabf} = \arg\min_\thetabf R^\star(\thetabf\smid p_{\tilde{\thetabf}})$ \cite{gribonval2017}\unmodifiedtext, the excess risk bound~\eqref{eq:excessriskbound} can be combined with the LRIP~\eqref{eq:LRIP} and the definition of $\tilde{\thetabf}$~\eqref{eq:SketchFitting2} as follows: 
\begin{align}
R^{\star}(\tilde{\thetabf} \smid  p_X) -R^{\star}(\finalrevisions{\thetabf^{\star}} \smid  p_X) 
&\leq 2 d(p_{\tilde{\thetabf}},p_X) \notag\\
&\leq 2 \Big[ d(p_{\tilde{\thetabf}},p_{\thetabf'}) 
              + d(p_{\thetabf'},p_X) \Big] \notag\\
&\leq 2 \Big[ C_0 \|\cA(p_{\tilde{\thetabf}})-\cA(p_{\thetabf'})\|  
              + d(p_{\thetabf'},p_X) \Big]  \notag\\
&\leq 2 \Big[ C_0 \|\cA(p_{\tilde{\thetabf}})-\empts\|  
+ C_0 \|\empts-\cA(p_{\thetabf'})\|  
              + d(p_{\thetabf'},p_X) \Big]  \notag\\    
&\leq 2 \Big[ 2C_0 \|\cA(p_{\thetabf'})-\empts\|  
              + d(p_{\thetabf'},p_X) \Big]  \notag\\    
&\leq 2 \Big[ 2C_0 \|\cA(p_X)-\empts\|  
              + 2C_0 \|\cA(p_{\thetabf'})-\cA(p_X)\|  
              + d(p_{\thetabf'},p_X) \Big]
               \label{eq:excessriskbound2}, 
\end{align}
for any distribution $p_{\thetabf'}$ in $\Sigma_{\Thetabf}$
(because $d(\cdot,\cdot)$ is a valid distance metric, \ie $d(p_1,p_2)\leq d(p_1,p_3)+d(p_3,p_2)~\forall p_3$).
One option is to choose $p_{\thetabf'}=p_{\thetabf^{\star}}$, in which case \modifiedtext the term \unmodifiedtext
$2C_{0}\|\cA(p_{\thetabf^{\star}})-\cA(p_X)\|+d(p_{\thetabf^{\star}},p_X)$ 
\modifiedtext in the upper bound \unmodifiedtext
reflects the excess risk due to modeling and 
$2C_0 \|\cA(p_X)-\empts\|$ reflects the excess risk due to sketching from a finite dataset $\ds$.
For a tighter bound, we could choose $p_{\thetabf'}$ as the distribution in $\Sigma_{\Thetabf}$ that minimizes the right side of~\eqref{eq:excessriskbound2}.
These approaches, and more refined variants, have been applied to analyze various learning tasks, in \eg \cite{gribonval2017,sheehan2019compressiveICA,sheehan2019compressive}. 

It should be emphasized that recovery guarantees based on the RIP~\eqref{eq:RIP} or LRIP~\eqref{eq:LRIP} hold \emph{even in the presence of measurement noise $\noise$ and/or modeling errors}.
In CS, measurement noise arises due to, \eg thermal noise or interference, while modeling errors arise when $\xbf_0$ is not truly $k$-sparse. 
Many sparse recovery techniques are provably robust  to such noise and modeling errors, \cf~\eqref{eq:l1-l2-instance-optimality} in~\ybref{yb:CS}.
In 
\modifiedtext compressive \unmodifiedtext
learning, measurement noise arises due to the finite cardinality of the dataset $\ds$ (recall~\eqref{eq:noise}), while modeling errors arise when $p_X\notin \Sigma_\Thetabf$.
For example, GMM-recovery guarantees can be established even when $p_X$ is not truly a GMM.

\subsection{Establishing the LRIP via kernels and the JL lemma}\label{sec:kernelJL}

\modifiedtext In light of the fact that the LRIP \eqref{eq:LRIP} yields statistical learning guarantees \eqref{eq:excessriskbound2} for compressive learning, a key question becomes: \unmodifiedtext
How can we choose the sketch dimension $m$ so that the LRIP~\eqref{eq:LRIP} holds?
Similar to how the RIP is proven in CS \modifiedtext (see, \eg~\cite[Lemma 9.33]{FouRau13}), refinements of a mathematical tool called the Johnson-Lindenstrauss (JL) lemma \cite{johnson1984extensions}
can be used to obtain a value of $m$ sufficient for the LRIP to hold \emph{with high probability on the draw of the random feature map $\fmap(\cdot)$.} \unmodifiedtext
We now summarize this approach.

\modifiedtext To begin, we establish \unmodifiedtext
connections to kernel methods (see~\ybref{yb:Kernel} for a brief review of kernels).
Some key observations are that any feature map explicitly defines a positive definite \emph{kernel}, 
and that the \emph{expectation} of 
\modifiedtext this kernel \unmodifiedtext
defines another useful positive definite kernel.
To see why, consider, for example, the RF feature map \eqref{eq:RFF}. 
First, for a fixed set of frequency vectors $\{\wbf_j\}_{j=1}^m$, we have
\begin{equation}\label{eq:shiftinvariant}
\langle \tfrac{1}{\sqrt{m}}\fmap(\xbf), \tfrac{1}{\sqrt{m}}\fmap(\xbf')\rangle 
= 
\frac{1}{m} \sum_{j=1}^{m} \exp(-\jmath 2\pi \wbf_{j}\tran(\xbf-\xbf')),\qquad \forall \vx,\vx'
\end{equation}
\modifiedtext where $\langle \cdot,\cdot \rangle$ is the standard Euclidean inner product in $\mathbb{R}^m$ or $\mathbb{C}^m$. \unmodifiedtext
The left-hand side defines a positive definite kernel according 
to the terminology reviewed in~\ybref{yb:Kernel}. 
This kernel is also ``shift-invariant,'' in that it depends only on the difference $\xbf-\xbf'$.
Next, imagine that the $d$-dimensional frequency vectors $\wbf_{j}$, which constitute the $m$ rows of the random matrix $\linop$, are drawn \iid from some probability distribution $p_{W}$. 
In the limit of large sketch dimension $m$, the law of large numbers combined with \eqref{eq:shiftinvariant} says 
\begin{equation}\label{eq:shiftinvariantasymptotic}
\langle \tfrac{1}{\sqrt{m}}\fmap(\xbf), \tfrac{1}{\sqrt{m}}\fmap(\xbf')\rangle 
 \stackrel{\wpone}{\rightarrow} \mathbb{E}_{W}  \exp(-\jmath 2\pi W\tran(\xbf-\xbf')).
 \end{equation}
The right hand side of \eqref{eq:shiftinvariantasymptotic} is the Fourier transform (FT) of the probability density $p_W$ evaluated at $\xbf-\xbf'$, \ie the characteristic function $\Psi_{p_W}(\xbf'-\xbf)$ using the notation from \eqref{eq:CF}.
Because $\Psi_{p_W}(\cdot)$ is the FT of a non-negative function, it is a so-called \emph{positive definite function}, which means that, for any $\{\xbf_i\}_{i=1}^n$, the $n\times n$ matrix $\Vec{\Psi}$ defined with elements $\Psi_{ij}=\Psi_{p_W}(\xbf_i-\xbf_j)$ for $1\leq i,j \leq n$ will be positive semi-definite.
Thus, if we construct a kernel as $\kappa(\xbf,\xbf'):=\Psi_{p_W}(\xbf'-\xbf)$, then it will be a positive definite kernel.
When $p_{W} = \cN(\Vec{0},\sigma_{w}^{2} \Vec{I}_d)$, this approach yields the 
familiar Gaussian kernel (a particular type of ``radial basis function''),
\ie $\kappa_{\sigma}(\vx,\vx') := \exp(- \|\vx -\vx'\|^2/2\sigma^2)$, here of width $\sigma = 1/\sigma_{w}$. 

More generally, by considering any parametric feature map of the form $\fmap(\vx\smid  \linop)$, where the parameter $\linop$ is drawn at random according to some probability distribution, one can define\footnote{In many papers, the $1/\sqrt{m}$ scaling in~\eqref{eq:shiftinvariant}-\eqref{eq:AverageKernel} is subsumed in the feature map $\fmap(\cdot)$.} the ``expected kernel''
\begin{equation}\label{eq:AverageKernel}
\kappa(\vx,\vx') := 
\mathbb{E}_{\linop} \left\langle  \tfrac{1}{\sqrt{m}} \fmap(\vx\smid  \linop), \tfrac{1}{\sqrt{m}} \fmap(\vx'\smid  \linop)\right\rangle,
\end{equation}
not to be confused with the ``mean kernel'' defined in \eqref{eq:meankernel}.
This setting includes
RF features~\eqref{eq:RFF} with \iid frequencies $\wbf_j$ (as above), or with a frequency matrix $\linop$ that includes structured blocks of rows, as we will soon discuss. 

\begin{highlight}[label={yb:Kernel}]{Kernel Methods and Kernel Embeddings of Probability Distributions}

Sketching shares connections with \emph{kernel methods} \cite{scholkopf2002learning}, a family of machine learning techniques that produce decisions or insights using a kernel function, $\kappa(\vx,\vx') \in \mathbb R$, which measures the ``similarity'' between $\vx$ and $\vx'$.
A kernel is said to be ``positive definite'' if the $n \times n$ matrix $\Kbf$, constructed with entries $\kappa(\vx_{i},\vx_{j})$ for $1 \leq i,j\leq n$, is positive \emph{semi}-definite for every possible $\{\vx_{i}\}_{i=1}^n$.
The celebrated ``kernel trick'' states that any positive definite kernel  \emph{implicitly} amounts to an inner product in some higher-dimensional (and potentially infinite-dimensional) feature space $\cH$, and vice-versa. 
That is, $\kappa(\vx,\vx') = \langle \bar{\fmap}(\vx), \bar{\fmap}(\vx') \rangle$ for some (not necessarily explicitly known) mapping $\bar{\fmap}(\cdot)$ 
from the signal space to $\cH$.
Any machine learning method that relies only on the evaluation of inner products---such as ridge regression, support-vector-machine classification, PCA \cite{vapnik2013nature}, and dictionary learning \cite{Rubinstein:2010aa}---can be ``kernelized'' by using a kernel in place of the inner product. 
Kernelizing a method is thus tantamount to applying that method in a transformed,
higher-dimensional feature space. 
In this way, more complex estimation and/or decision functions can be implemented.

Given a positive definite kernel $\kappa(\vx,\vx')$ operating on signals $\vx$ and $\vx'$ in some set, it is possible to ``lift'' $\kappa(\cdot,\cdot)$ to a positive definite kernel \emph{operating on probability distributions} over this set by defining the so-called \emph{mean kernel} 
\begin{equation}\label{eq:meankernel}
k(p,q) := \mathbb{E}_{X \sim p, X' \sim q} [\kappa(X,X')].
\end{equation}
This defines an embedding of probability distributions into a kernel space, which is analogous to the finite-dimensional embedding $\cA(p)$ of probability distribution $p$ from~\eqref{eq:ProjDef}.
The \emph{maximum mean discrepancy} (MMD) \cite{Gretton2007, Sriperumbudur2010}
\begin{equation}
\MMD(p,q) := \sqrt{k(p,p) + k(q,q) - 2k(p,q)}
\label{eq:MMD}
\end{equation}
is the Euclidean metric naturally induced by the mean kernel. 
It is analogous to the Euclidean distance between sketches, $\|\cA(p)-\cA(q)\|$.
The MMD, originally introduced in the context of two-sample hypothesis testing \cite{Gretton2007}, is now well-known in machine learning. 
When the MMD behaves as a true metric, \ie when $\MMD(p,q)=0 \Leftrightarrow p=q$, the mean kernel $k(\cdot,\cdot)$ is said to be ``characteristic''. 
In $\mathbb{R}^d$, many classic kernels $\kappa(\cdot,\cdot)$, such as the Gaussian and Laplace kernels, yield characteristic mean kernels \cite{Sriperumbudur2010}.
\end{highlight}

Now that the kernel connections have been established, we return to our original objective of \emph{understanding when the LRIP~\eqref{eq:LRIP} holds}. 
Importantly, the law of large numbers shows that, in the limit of large sketch dimension $m$, the right side of \eqref{eq:LRIP} is related to the maximum mean discrepancy (MMD) from \eqref{eq:MMD}, which is a kernel-based distance between distributions (see~\ybref{yb:Kernel}).
Indeed, for arbitrary probability distributions $p$ and $q$, we have
\begin{equation}
\label{eq:normMMD}
\lim_{m \to \infty}\frac{1}{\sqrt{m}} \|\cA(p)-\cA(q)\| \stackrel{\wpone}{=} \MMD(p,q) ,
\end{equation}
\finalrevisions{where the maximum mean discrepancy is implicitly the one obtained from the kernel used to build the sketching operator~$\cA$.}
Thus, 
\modifiedtext when the LRIP~\eqref{eq:LRIP} holds, it must also be true that, for sufficiently large sketch dimension $m$,\unmodifiedtext
\begin{equation}\label{eq:KernelLRIP}
d(p_{\thetabf},p_{\thetabf'}) \leq C'_{0} \MMD(p_{\thetabf},p_{\thetabf'}),\qquad \text{for all}\ p_{\thetabf},p_{\thetabf'} \in \Sigma_{\Thetabf},
\end{equation}
where $C'_{0}$ is a positive constant. Property~\eqref{eq:KernelLRIP} connects two different metrics on probability distributions: the left hand side of \eqref{eq:KernelLRIP} is defined by the learning task (recall~\eqref{eq:distance}), while the right hand side of \eqref{eq:KernelLRIP} is defined by the expected kernel $\kappa$, \modifiedtext from\unmodifiedtext~\eqref{eq:AverageKernel}, associated with the randomized feature map.
Note that \eqref{eq:KernelLRIP} is a deterministic property; it does not depend on the draw of the randomized feature map, unlike the LRIP \eqref{eq:LRIP}.
In the literature,~\eqref{eq:KernelLRIP} is called the ``kernel-LRIP'' \cite{gribonval2017} because it is a kernel-based analog to the LRIP.
Being deterministic, the expected kernel is \modifiedtext often \unmodifiedtext easier to manipulate than the random feature map, and thus eases the proof of the kernel-LRIP. 
This is important because, if the kernel-LRIP \eqref{eq:KernelLRIP} holds, then, using arguments based on the JL lemma, one can also establish \cite{gribonval2017} the LRIP~\eqref{eq:LRIP}, as we show below.

The JL lemma \cite{johnson1984extensions} is a precursor of the restricted isometry property that is specialized to finite sets~\eqref{eq:RIP}. It states that, given $N$ arbitrary $d$-dimensional vectors $\xbf_{i}$, there exists an $m \times d$ matrix $\Abf$, with $m$ on the order of $\log N$, such that $\|\Abf \xbf_{i}-\Abf \xbf_{j}\| \approx \|\xbf_{i}-\xbf_{j}\|$ for all $i,j$. 

To illustrate how the JL lemma is useful in the context of 
\modifiedtext compressive \unmodifiedtext
learning,
let us momentarily restrict our attention to a learning problem where the collection $\Sigma_{\Thetabf}$ of parametric distributions is of \emph{finite cardinality}, 
noting that we can extend this approach to continuous families of parametric distributions through discretization arguments involving the notion of covering numbers \cite[Appendix C]{FouRau13}, as described below. 
As a concrete example, let us consider a discretized variant of 
\modifiedtext compressive \unmodifiedtext 
GMM using the RF feature map~\eqref{eq:RFF}. As in \cite{keriven2017b,gribonval2017}, we assume that the $d$-dimensional frequency vectors $\wbf_{j}$ are drawn \iid from the normal distribution $\cN(\Vec{0},\sigma_{w}^{2} \Vec{I}_d)$, and we consider learning a mixture of Gaussian components $p_{\thetabf_\ell} = \cN(\mubf_\ell, \Ibf)$ for $\ell=1,\dots,k$, with equal weights $\alpha_{\ell}=1/k$ for each $\ell$, where the means $\mubf_{\ell}$ are assumed to be bounded, separated, and discretized on a regular grid. 
In this setting, there exists a finite number, $N$, of possible parametric mixture distributions $p_{\thetabf} \in \Sigma_{\Thetabf}$. \modifiedtext As long as the MMD is a true metric (as defined in \ybref{yb:Kernel}), the ratio $d(p_{\thetabf},p_{\thetabf'}) / \MMD(p_{\thetabf},p_{\thetabf'})$ is finite for $p_{\thetabf} \neq p_{\thetabf'}$, hence whenever \unmodifiedtext
$\Sigma_{\Thetabf}$ is a finite collection, as in this concrete example, the kernel-LRIP~\eqref{eq:KernelLRIP} holds (for a sufficiently large $C_0'$).
Moreover, specializing to an arbitrary pair of mixtures $p_{\thetabf},p_{\thetabf'} \in \Sigma_{\Thetabf}$ and a given sketch dimension $m$, refinements of~\eqref{eq:normMMD} (using \emph{measure concentration}) enable one to show that the lower bound 
\begin{equation}
\label{eq:LowerBound}
\frac{1}{\sqrt{2}} {\MMD} (p_{\thetabf},p_{\thetabf'})
\leq \frac{1}{\sqrt{m}} \|\cA(p_{\thetabf})-\cA(p_{\thetabf'})\|
\end{equation}
holds with probability at least $1-\exp(-c_{0}m)$, where $c_{0}$ is a positive \emph{concentration constant}. 
\finalrevisions{Since there are only $N^{2}$ pairs of parametric mixtures $p_{\thetabf},p_{\thetabf'} \in \Sigma_{\Thetabf}$, the lower bound~\eqref{eq:LowerBound} is valid \emph{uniformly} for all of them, except with probability at most $N^{2} \exp(-c_{0}m)$. 
This failure probability can be made smaller than any $\epsilon>0$ by choosing $m$ larger than $2 c_{0}^{-1}\log (N/\sqrt{\epsilon})$. 
}
Combining \eqref{eq:KernelLRIP} and \eqref{eq:LowerBound} using a union bound,
it can be shown \cite{gribonval2017} that the LRIP~\eqref{eq:LRIP} holds with high probability when $C_{0}$ is on the order of $C'_{0}\sqrt{m}$ and the sketch dimension $m$ grows logarithmically with $N$, the number of parametric distributions in the finite set $\Sigma_{\Thetabf}$. 

For infinite collections $\Sigma_{\Thetabf}$, proving \modifiedtext the kernel-LRIP\unmodifiedtext~\eqref{eq:KernelLRIP}, and eventually the LRIP~\eqref{eq:LRIP}, is more technical and can require some additional assumptions.
As an example, for 
\modifiedtext compressive \unmodifiedtext 
clustering with RF features,
it can be proven that there is a constant $C'_{0}$ such that~\eqref{eq:KernelLRIP} holds, provided
that the centroids are sufficiently separated and bounded \cite{gribonval2017}.
The JL lemma can be extended by refining techniques used to establish the RIP~\eqref{eq:RIP}. 
\modifiedtext The main idea is to first prove the LRIP on some finite collection $\Sigma' \subset \Sigma_{\Thetabf}$ of $N$ probability distributions, and then to extrapolate it (with slightly worse constants) to $\Sigma_{\Thetabf}$. \unmodifiedtext 
Technically, this involves the notion of covering numbers, and the cardinality $N=|\Sigma'|$ is typically exponential in the number of parameters needed to describe $\Sigma_{\Thetabf}$. For example, for GMM, one needs $kd$ parameters to describe the means $\mubf_{\ell} \in \mathbb{R}^{d}$, $1 \leq \ell \leq k$ of the mixture $p_{\thetabf} = \sum_{\ell=1}^{k} \frac{1}{k} \cN(\mubf_{\ell},\Ibf)$, and $\log N$ essentially depends linearly on $kd$.
The dimension $m$ of the sketch for which the LRIP holds with high probability is thus on the order of $kd/c_{0}$, up to some additional factors due to the proof technique \cite{gribonval2017}.

Empirical studies of 
\modifiedtext compressive \unmodifiedtext 
clustering \cite{keriven2017a,byrne2019sketched} and 
\modifiedtext compressive \unmodifiedtext 
GMM \cite{keriven2017b} suggest that a sketch dimension $m$ on the order of $kd$ (the number of parameters in these settings) is sufficient to yield accurate learning performance. 
The best known bounds on provably good sketch dimensions \cite{gribonval2017} remain pessimistic compared to these empirically validated sketch dimensions.  
This is most likely related to sub-optimal bounds for the concentration constant $c_{0}$ and/or shortcomings in the techniques used to extend the LRIP from a finite collection $\Sigma'$ to an infinite collection $\Sigma_{\Thetabf}$.

\subsection{The challenge of designing a feature map given a learning task}
\label{sec:design}

In the existing literature, the LRIP has been established \cite{gribonval2017} for randomized feature maps $\fmap(\cdot)$ (\eg random Fourier features, random quadratic features) that mimic related constructions from compressive sensing, developed either for sparse-vector recovery or low-rank matrix recovery. 

When sketching with random Fourier features (\eg for 
\modifiedtext compressive \unmodifiedtext 
clustering and 
\modifiedtext compressive \unmodifiedtext 
GMM), the main design choice for $\fmap(\cdot)$ is the distribution from which to draw the random frequencies $\wbf_j$ (\ie the rows of $\linop$ in \eqref{eq:RFF}). 
In light of the connections to shift-invariant kernels (recall~\eqref{eq:shiftinvariant}), this design task is a particular instance of the difficult problem of kernel design \cite[Sec. 4.4.5]{scholkopf2002learning}.

For example, when the rows of $\linop$ are drawn \iid $\cN(\Vec{0},\sigma_{w}^2\Ibf)$, the choice of the variance $\sigma_{w}^2$ determines the choice of the width $\sigma = 1/\sigma_{w}$ of the corresponding Gaussian kernel $\kappa_{\sigma}(\cdot,\cdot)$.
Indeed, from a signal-processing standpoint, the corresponding mean kernel $k_{\sigma}(\cdot,\cdot)$ (recall~\eqref{eq:meankernel}) acts to low-pass filter the underlying data distributions. 
To see why, observe that 
$\kappa_{\sigma}(\vx,\vx') = g_{\sigma}(\vx-\vx')$ with $g_{\sigma}(\vx) := e^{-\|\vx\|^{2}/2\sigma^{2}}$, and so
\begin{align}
k_{\sigma}(p,q)
&= \bb E_{X \sim p, X' \sim q} [\kappa_{\sigma} (X,X')] 
= \iint e^{-\|X-X'\|^{2}/2\sigma^{2}} \dif p(X) \dif q(X') \\
&= \langle g_{\sigma} \star p, q\rangle_{L^2} 
= \langle g_{\bar \sigma} \star p, g_{\bar \sigma} \star q\rangle_{L^2} , \label{eq:GaussianMeanKernel}
\end{align}
where $\star$ denotes convolution and $\bar \sigma = \sigma/\sqrt{2}$. 
Hence, the associated MMD~\eqref{eq:MMD} satisfies
$\MMD(p,q) = \|g_{\bar \sigma} \star p- g_{\bar \sigma} \star q\|_{L^{2}}$.
Recall that learning-from-a-sketch is often performed by minimizing a ``sketch matching'' cost $\|\empts-\cA(p_{\thetabf})\|^{2} = \|\cA(\hat{p}_{\ds}) - \cA(p_\thetabf)\|^{2}$, as in~\eqref{eq:SketchFitting}, where $\hat{p}_{\ds}$ denotes the empirical distribution of the data $\ds$. 
In the limit of large sketch dimension $m$, this cost compares the smoothed versions of the probability distributions $\hat{p}_{\ds}$ and $p_\thetabf$, since $\tfrac{1}{m}\|\empts-\cA(p_{\thetabf})\|^{2}  \approx \big \|g_{\bar{\sigma}} \star  \hat{p}_{\ds} - g_{\bar{\sigma}} \star  p_\thetabf \big\|_{L_2}^{2}$. 
Similarly, when using a greedy algorithm to learn a mixture model (or cluster centroids) from a sketch, the normalized inner product $\tfrac{1}{m} \langle \empts, \cA(p_{\thetabf_{\ell}})\rangle$ approximates the correlation $\langle g_{\bar{\sigma}} \star \hat{p}_{\ds},g_{\bar{\sigma}} \star p_{\thetabf_{\ell}} \rangle_{L_2}$ between the low-passed versions of the empirical data distribution $\hat{p}_{\ds}$ and the candidate mixture component $p_{\thetabf_{\ell}}$, respectively. 

\begin{figure}[!t]
        \centering
        \includegraphics[width=0.8\linewidth]{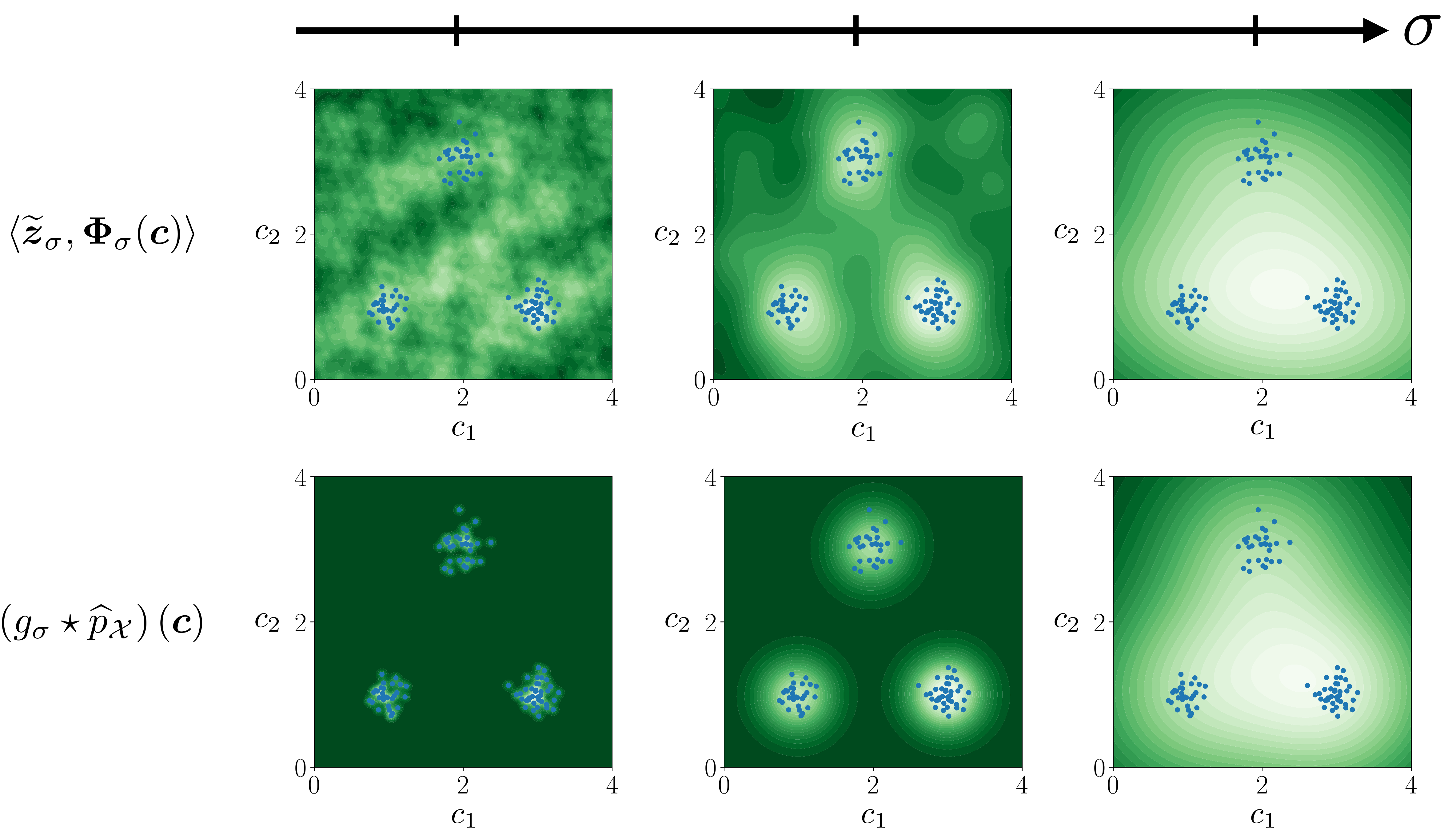}
        \label{fig:tmpexp21}
  \caption{Criterion $\langle\empts_{\sigma},\fmap_{\sigma}(\vc)\rangle$ used by greedy parameter estimation algorithms in 
  \modifiedtext compressive \unmodifiedtext 
clustering with random Fourier features (top row) versus its expected value $(g_{\sigma} \star \hat{p}_{\ds})(\cbf)$ (bottom row) as a function of the centroid hypothesis location $\bs c = [c_1, c_2]\tran$. The dataset (in blue) consists of $n= 100$ points drawn according to a mixture of $k=3$ isotropic Gaussians. The frequencies $\wbf_{j}$ used to define the feature map $\fmap_{\sigma}(\cdot) = \exp(-\jmath 2\pi \linop \cdot)$ are drawn according to a standard Gaussian $\cN(\Vec{0},\sigma_{w}^{2}\Ibf)$ with $\sigma_{w}=1/\sigma$. The sketch $\empts_{\sigma}$ is computed with the feature map $\fmap_{\sigma}(\cdot)$.  With $\sigma^2=\frac{1}{500}$ (left), there is insufficient smoothing, and the criterion displays many spurious local maxima; with $\sigma^2=\frac{1}{10}$ (middle), there is appropriate smoothing, and local maxima are in good correspondence with the true cluster centers; with $\sigma^2=1$ (right), there is over-smoothing, and the criterion displays only a single maximum.  
  }
  \label{fig:lowpass}
\end{figure}

This latter idea is illustrated for 
\modifiedtext compressive \unmodifiedtext 
clustering in \autoref{fig:lowpass}. 
There, since the mixture component associated to a candidate centroid $\cbf$ is the Dirac $p_{\cbf}(\vx)= \delta(\vx-\cbf)$ (recall the discussion before \eqref{eq:learnkmeansextended}), we have that $\langle \empts_\sigma, \cA_\sigma(p_{\cbf})\rangle = \langle \empts_\sigma,\fmap_\sigma(\cbf)\rangle$, where the dependence on the kernel width $\sigma=1/\sigma_w$ has been made explicit.
Meanwhile,
\begin{equation}\label{eq:Parzen}
\langle g_{\bar{\sigma}} \star \hat{p}_{\ds},g_{\bar{\sigma}} \star p_{\cbf} \rangle_{L_2} 
= (g_{\sigma} \star \hat{p}_{\ds})(\cbf)
= \frac{1}{n} \sum_{i=1}^{n} g_{\sigma}(\cbf-\xbf_{i}).
\end{equation}
In \eqref{eq:Parzen}, we recognize a Parzen window density estimator \cite{pan2008parzen}, whose computation for a given $\cbf$ requires access to the $n$ training samples $\xbf_{i}$. 
In contrast, its surrogate $\langle \empts_\sigma, \fmap_\sigma(\cbf)\rangle$ only requires access to the $m$-dimensional sketch $\empts$. 
In a large-scale setting, this can save huge amounts of memory and computation. 
As can be seen when comparing the top and bottom rows in \autoref{fig:lowpass}, $\langle \empts_{\sigma},\fmap_{\sigma}(\cbf) \rangle$ well approximates $(g_{\sigma} \star \hat{p}_{\ds})(\cbf)$ for sufficiently large kernel width $\sigma$.
By comparing the different columns of \autoref{fig:lowpass}, it can also be seen that the kernel width $\sigma$ should be chosen compatible with the cluster width and separation. 
This choice involves a tradeoff between smoothing the unwanted gaps between data samples and over-smoothing the (desired) gaps between clusters. 
Existing theory \cite{gribonval2017} identifies sufficient conditions on the choice of $\sigma$ related to the number $k$ of candidate clusters and their minimum separation. 

Although \iid Gaussian frequencies $\wbf_{j}$ were used with the RF map above, one may also consider the use of \iid non-Gaussian frequencies.
Such designs, as proposed in \cite{keriven2017b}, can yield improved empirical behavior.
As we will see in the next section, it is also possible to 
deviate from the RF map with the goal of improving computational efficiency.
Such constructions also yield non-Gaussian expected kernels~\eqref{eq:AverageKernel}. 
Although they work well in practice, there is currently no proof that these latter kernels satisfy the kernel-LRIP.

While existing theory focuses on proving the LRIP for a given random feature map and learning task, an important open question is: 
How should one \emph{design the feature map} to best match a given learning task?
In particular, can we design a random feature map that satisfies the LRIP~\eqref{eq:LRIP} for a given learning task defined by a loss function $L(\thetabf \smid \xbf_i)$ and embodied by a task-driven distance~\eqref{eq:distance}? 
A promising \modifiedtext yet still challenging \unmodifiedtext avenue would be to first identify a positive definite kernel $\kappa_{0}(\vx,\vx')$ for which the corresponding MMD satisfies the kernel-LRIP~\eqref{eq:KernelLRIP}, and then use Bochner's theorem or Mercer's theorem (see~\ybref{yb:Mercer}) to design a random feature map $\fmap(\cdot)$ whose expected kernel~\eqref{eq:AverageKernel} is precisely $\kappa_{0}$.

\begin{highlight}[label={yb:Mercer}]{Approximating kernel methods with random feature maps}

While each random feature map $\fmap(\cdot)$ implicitly defines a positive definite kernel by taking the expectation~\eqref{eq:AverageKernel}, the converse is also true.

For shift-invariant kernels, \ie kernels for which $\kappa(\vx,\vx') = \kappa(\vx-\vx',0)$ only depends on the difference $\vx-\vx'$ (such as the Gaussian kernel), this is a consequence of Bochner's theorem~\cite{Rahimi2007}, which states that if $\kappa$ is a positive definite kernel such that $\kappa(\vx,\vx)=1$ for all $\vx$,  then its Fourier transform yields a probability distribution $p_W(\wbf) = \int \kappa(\vx,0) \exp(-\jmath 2\pi \wbf\tran\vx) \dif\vx$. 
Conversely, the kernel can be obtained by the inverse Fourier transform, which can also be phrased as an expectation:
 \[
 \kappa(\vx,\vx') = \int \exp\left(\jmath 2\pi \wbf\tran (\vx-\vx')\right) p_{W}(\wbf) \dif\wbf
 = \bb E_{W \sim p_{W}}  \exp\left(\jmath 2\pi W\tran (\vx-\vx')\right).
 \]
 Hence, drawing \iid frequency vectors $\wbf_{j}$ according to $p_{W}$ yields an RF feature map~\eqref{eq:RFF} whose expected kernel is precisely $\kappa$.
For instance, a Laplace kernel can be approximated if the rows of $\linop$ are drawn \iid from the Cauchy distribution~\cite{boufounos2017representation}. 

More generally, under mild assumptions on a positive definite kernel $\kappa$, one can invoke Mercer's theorem \cite{Bach2017} to similarly show the existence of a random feature map whose expected kernel, in the sense of~\eqref{eq:AverageKernel}, matches $\kappa$. 
\end{highlight}

\section{Sketching with reduced computational resources} 
\label{sec:structured}

The computational cost of sketching via~\eqref{eq:sketch-def} is heavily dependent on the feature map $\fmap(\cdot)$. 
The computational cost of parameter estimation via~\eqref{eq:learngeneric} is also heavily dependent on $\fmap(\cdot)$, since it often involves iterative application of $\fmap(\cdot)$.

Often, the feature map is constructed as a randomized linear operation followed by a componentwise non-linear operation, \ie 
\begin{equation}
\label{eq:GenericSketching}
\fmap(\vx)=\varrho(\linop\vx),
\end{equation}
where $\linop$ 
is a (randomly drawn) matrix of size $m \times d$ and $\varrho(\cdot)$ applies a scalar non-linear function identically to each element of the vector $\linop\vx$. 
For example, the feature maps described earlier for 
\modifiedtext compressive \unmodifiedtext
PCA, GMM, and clustering all have this form.
Reducing the computational cost of each stage has been the goal of several studies. 
For example, using a fast transform for $\linop$ drastically reduces the memory and computational complexity demands relative to an explicit matrix.
Also, quantized versions of the nonlinearity $\varrho(\cdot)$ are much more easily implemented in hardware than, say, the complex exponential nonlinearity $\varrho_{\cexp}(\cdot) := \exp(-\jmath 2\pi \: \cdot)$ used in the RF map~\eqref{eq:RFF}.
We discuss such constructions of $\linop$ and $\varrho(\cdot)$ below.

\subsection{Sketching with structured random matrices} 
\label{sec:matrices}

In most of our previous examples, we constructed $\linop$ by drawing its $m$ rows \iid from the normal distribution $\cN(\Vec{0},\sigma_w^2\Ibf_d)$ with some variance $\sigma_w^2>0$.
Recall that, with the RF map $\varrho_{\cexp}(\cdot)$, the rows of $\linop$ correspond to the ($d$-dimensional) frequencies used when sampling the 
 Fourier transform of the (empirical) data distribution.
In any case, when $\linop$ is an explicit matrix, the computational complexity of computing $\fmap(\vx)$ of the form~\eqref{eq:GenericSketching} is dominated by the matrix-vector product $\linop\vx$. 
Thus, it is on the order of $md$, which is also the order of the memory needed to store $\linop$.

\begin{figure}
        \newcommand{\tikzb}[1]{{\color{black}$\bs B\tran_#1$}}
  \begin{center}
        \ifrecompiletikz
			\input{fig-structured-matrices}
		\else
        	\includegraphics[width=0.9\textwidth]{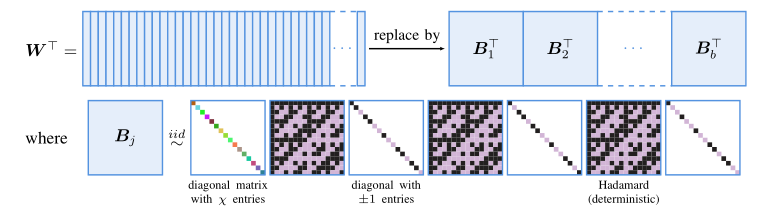}
		\fi
  \end{center}
  \caption{\label{f:structure}Structured random matrix design from \cite{yu2016}. Each block is a composition of several Hadamard and diagonal matrices. For convenience, we draw $\linop\tran$ instead of $\linop$.}
\end{figure}

As an alternative to these approaches, it has been suggested to construct $\linop$ as a structured random matrix with a fast implementation that mimics an \iid Gaussian matrix.
Multiple ways of accomplishing this goal have been proposed in the literature.
We focus on the approach suggested in \cite{yu2016}, which was successfully applied to 
\modifiedtext compressive \unmodifiedtext 
learning in~\cite{chatalic2018}.
There, the idea is to construct $\linop$ as a vertical concatenation of $b=\lceil m/d \rceil$ blocks $\{\bs B_j\}_{j=1}^b$, each of size $d\times d$.
These blocks have the form $\bs B_j=\bs D_j^{(0)}\bs H\bs D_j^{(1)}\bs H\bs D_j^{(2)}\bs H\bs D_j^{(3)}$, where $\bs H$ is the Walsh-Hadamard matrix and $\bs D_j^{(k)}$ are random diagonal matrices.
In particular, the diagonal elements of $\bs D_j^{(1)}$, $\bs D_j^{(2)}$, and $\bs D_j^{(3)}$ are drawn \iid from the uniform distribution over $\{-1,1\}$, and the diagonal elements of $\bs D_j^{(0)}$ are drawn \iid from the $\chi$ distribution with $d$ degrees-of-freedom, which is the distribution of the norm of a $d$-variate Gaussian vector. 
This construction is depicted in Fig.~\ref{f:structure}. 
The fast Walsh-Hadamard transform offers an order $d\log d$-complexity implementation of the matrix-vector multiplication $\bs H\vx$ and prevents the need to explicitly store $\bs H$.
With this structured and fast incarnation of $\linop$, the sketching complexity shrinks from order $md$ to order $m\log d$.
Moreover, since only the diagonal matrices need to be stored, the storage cost shrinks from order $md$ to order $m$. 

\subsection{Sketching with quantized contributions}

With the RF map~\eqref{eq:RFF}, which is commonly used in 
\modifiedtext compressive \unmodifiedtext 
clustering and GMM, the non-linear operation $\varrho(\cdot)$ in~\eqref{eq:GenericSketching} becomes $\varrho_{\cexp}(\cdot) := \exp(-\jmath 2\pi \: \cdot)$.
Since implementing $\varrho_{\cexp}(\cdot)$ with high accuracy is somewhat costly and not amenable to easy hardware acceleration, one might consider quantizing it.
For example, dropping the imaginary part of $\varrho_{\cexp}(\cdot)$ and quantizing the real part to one bit of precision yields the $2\pi$-periodic function $\varrho_{q}(\cdot):=\sign(\cos(2\pi\cdot))$ 
which is simply a ``square wave''.

\begin{figure}[!t]
        \centering
  \includegraphics[width=\textwidth]{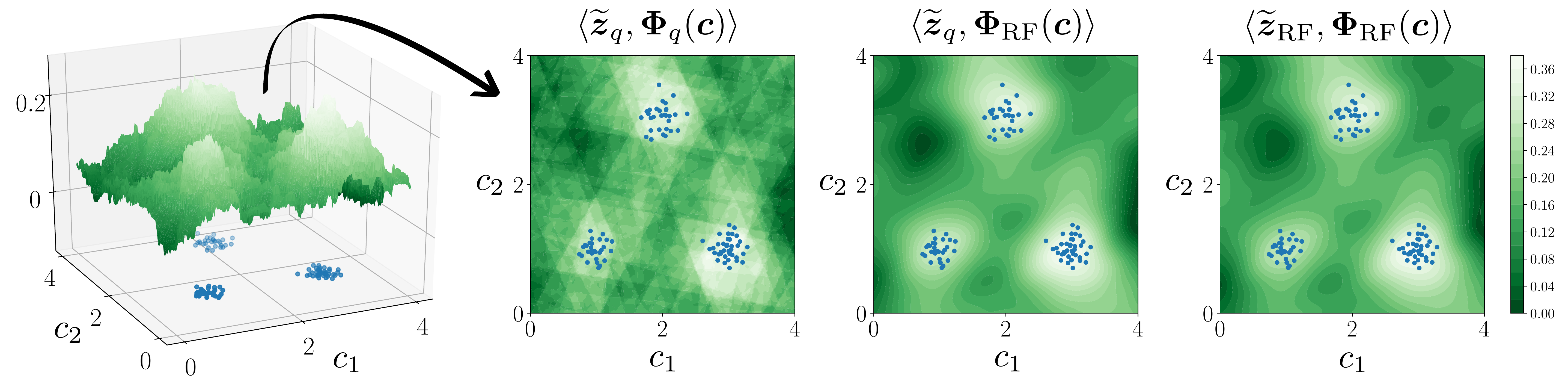}
  \caption{
  Criterion $\langle\empts,\fmap(\vc)\rangle$ versus centroid hypothesis $\bs c = [c_1, c_2]\tran$ used by greedy parameter estimation algorithms in 
  \modifiedtext compressive \unmodifiedtext 
clustering with $k=3$ and a $2$D dataset (in blue), with and without quantization of the sketch $\empts$ and/or feature map $\fmap$. The surface plot on the left highlights the irregularity of $\cbf \mapsto \langle \empts_{q},\fmap_{q}(\cbf)\rangle$.
  }
  \label{fig:univ-quant}
\end{figure}

To alleviate the effects of quantization, one can apply dithering \cite{gray1998quantization} to the input.
In this case, the feature map becomes $\fmap_q(\vx) = \varrho_{q}(\linop\vx + \bs\xi) \in \{-1,+1\}^m$, with \iid dither components $\xi_j$ drawn uniformly over $[0,1)$.
The effect of dithering is to make the quantized $\fmap_q$ behave similarly to non-quantized $\fmap_{\cexp}$ \emph{on average}.
For instance, it was shown in~\cite{schellekens2018a} that for each $\linop$, $\bs x$, $\bs x'$, and $\bs \xi$,
\begin{equation}
\label{eq:quantizedKernel}
\scp{\fmap_q(\bs x)}{\fmap_{\cexp}(\bs x')} \approx \bE_{\bs\xi}\scp{\fmap_q(\bs x)}{\fmap_{\cexp}(\bs x')} = c \scp{\fmap_{\cexp}(\bs x)}{\fmap_{\cexp}(\bs x')},
\end{equation}
where $c$ is a constant. 
The approximation above is accurate for a typical draw of the $m$-dimensional dither vector $\bs \xi$ when the sketch dimension $m$ is large enough~\cite{schellekens2020breaking}.
Note that, when this dithered quantizer is used to compute a sketch $\empts_q := \frac{1}{n} \sum_{i=1}^n \fmap_q(\vx_i) = \frac{1}{n} \sum_{i=1}^n \varrho_{q}(\linop\vx_i+\bs\xi)$, it is important to use the same dither realization $\bs\xi$ for all samples $i$.

The question then arises: when estimating parameters $\thetabf$ via~\eqref{eq:SketchFitting2}, how should we account for quantization in the sketch?
Simply replacing the $\cA(p_{\thetabf_{\ell}})$ term with $\cA_{q}(p_{\thetabf_{\ell}}) := \Exp_{X \sim p_{\thetabf_{\ell}}}[\Phibf_{q}(X)]$ may sound appealing. 
For example, with 
\modifiedtext compressive \unmodifiedtext
GMM~\eqref{eq:learnmixture} or 
\modifiedtext compressive \unmodifiedtext 
clustering~\eqref{eq:learnkmeans}, this would mean minimizing $\big\| \empts_q - \sum_{\ell=1}^k \alpha_\ell\, \cA_q(p_{\thetabf_{\ell}}) \big\|^{2}$. 
However, the optimization problem~\eqref{eq:SketchFitting2} would become more challenging, as suggested by comparing the left two panels to the right two panels in Fig.~\ref{fig:univ-quant}, which plots the centroid-selection criterion $\langle \empts, \fmap(\cbf)\rangle = \langle \empts, \cA(p_{\cbf})\rangle$ used by greedy algorithms.
Also, this approach would not inherit the theoretical guarantees that were carefully established using the LRIP~\eqref{eq:LRIP}, which does not easily translate to the quantized case. 

Instead, we suggest to use $\cA_{\cexp}(p_{\thetabf_{\ell}}) := \Exp_{X \sim p_{\thetabf_{\ell}}}[\fmap_{\cexp}(X)]$ for the $\cA(p_{\thetabf_{\ell}})$ term in 
\eqref{eq:learnmixture}, and to re-scale $\empts'_q = \empts_q/c$ with the constant $c$ from~\eqref{eq:quantizedKernel}.
Indeed, thanks to~\eqref{eq:quantizedKernel}, the resulting cost function $C(\thetabf|\empts'_q)=\big\| \empts'_q - \sum_{\ell=1}^k \alpha_\ell\, \cA_{\cexp}(p_{\thetabf_{\ell}}) \big\|^2$ is, in expectation over $\bs\xi$, \emph{exactly the same} (up a constant additive bias term) as the non-quantized cost function $C(\thetabf|\empts_{\cexp})$~\cite{schellekens2018a}.
The similarity between $C(\thetabf|\empts_q')$ and $C(\thetabf|\empts_{\cexp})$, even for a single realization of the $m$-dimensional dither vector $\bs\xi$, is suggested by comparing the 
right two panels in Fig.~\ref{fig:univ-quant}. 

Empirical results \cite{schellekens2018a} suggest that, when the sketch dimension is inflated by about 25\%, this quantized 
\modifiedtext compressive \unmodifiedtext 
learning procedure yields the same performance as the non-quantized procedure. 
Moreover, accurate probabilistic bounds for approximation~\eqref{eq:quantizedKernel}, established in~\cite{schellekens2020breaking}, allow one to extend the theoretical 
\modifiedtext compressive \unmodifiedtext 
learning guarantees in~\eqref{eq:excessriskbound2} to this new cost function.

\begin{highlight}[label={yb:Privacy}]{Potential of Sketching for Privacy-Aware Learning and Alternative Approaches}
Given a dataset, privacy preservation can be achieved by asking a trusted dataholder to 
corrupt
all queries of the dataset \cite{fung2010privacypreserving,dwork2014b}
in a controlled manner.
There are, however, challenges to this so-called \textit{interactive} approach.
For example, because
the privacy-preserving effects of this corruption can often be diminished through the mining of multiple query responses
(especially if the queries are adaptive), the per-query corruption levels must be designed with the type and total number of queries in mind.
These corruption levels are often designed 
using a so-called ``privacy budget,'' which is expended over multiple queries in order to meet an overall privacy level.
Once the entire privacy budget has been used up, the data can no longer be accessed by a given datauser.
Also, the dataholder must ensure that responses to \emph{different} datausers cannot be combined in a way that circumvents the intended privacy preservation. 

In contrast, the \emph{non-interactive} 
approach \cite{fung2010privacypreserving,dwork2014b} is to publish an intermediate privacy-preserving synopsis of the dataset, to which the public is allowed unlimited access.
For example, with a low-dimensional dataset, one could publish a privacy-preserving histogram of the data~\cite{qardaji2013differentially}, from which aggregate statistics could be subsequently extracted.
The non-interactive approach is attractive for several reasons.
For example, there is no need to formulate nor allocate a privacy budget; it is sufficient to set an overall privacy level.
Also, there is no need to worry about datausers sharing/combining data.

By adding noise to a sketch of the form \eqref{eq:sketch-def}, one can easily generate a privacy preserving synopsis of a dataset. 
By construction, such sketches capture the global statistics of the dataset $\ds=\{\vx_i\}_{i=1}^n$ while being relatively insensitive to each individual data sample $\vx_i$, especially when the sample cardinality $n$ is large.
Also, when the original data is distributed across multiple devices, 
a privacy preserving global sketch can be constructed by first locally sketching at each device and then averaging those local sketches at a fusion center, 
as illustrated in \ybref{yb:audio}. 
In this scenario, the local sketches will themselves be privacy preserving, which alleviates concerns about privacy leaks during data fusion.

\end{highlight}

\section{Privacy preservation} 

In addition to its efficient use of computational resources, 
sketching is a promising tool for privacy-preserving machine learning.
In numerous applications, such as when working with medical records, online surveys, or measurements coming from personal devices, data samples contain sensitive personal information and data providers ask that individuals' contributions to the dataset remain private, \ie not publicly discoverable.
Learning from such data collections while protecting the privacy of individual contributors has become a crucial challenge~\cite{fung2010privacypreserving,dwork2014b,sarwate2013signal}.

A common way to preserve privacy is to have a trusted dataholder (or ``curator'') corrupt the
response to each query of the dataset~\cite{fung2010privacypreserving} in a controlled manner.
A query may ask for something as simple as counting the number of times a given event occurred, or it may ask for more sophisticated information that requires the dataholder to run an inference algorithm.
As the corruption becomes more significant,
the privacy guarantee gets stronger, but the quality of the response to the query (called the ``utility'') degrades. 
This can be conceptualized by a \emph{privacy-utility tradeoff} \cite{fung2010privacypreserving,sarwate2013signal}.

When sketching a dataset via \eqref{eq:sketch-def}, 
there is a very simple way to preserve privacy in any subsequent learning task: simply add \iid noise (with appropriate distribution) to the sketch.
The privacy level can be adjusted by changing the variance of that noise, as described below. 
This one-time approach to 
privacy preservation
(more generally known as \emph{privacy-preserving data publishing} \cite{fung2010privacypreserving}) has several benefits over the query-based approach to 
privacy preservation
discussed in the previous paragraph.
(See also Box~\ref{yb:Privacy}.) 

\subsection{Sketching with differential privacy guarantees}
\label{s:dp}

\emph{Differential privacy}~\cite{dwork2014b} 
is a standard framework for privacy preservation that has a precise mathematical definition and is well-known in machine learning and signal processing (\eg \cite{sarwate2013signal}).
When a given (randomized) learning pipeline is differentially private, its output depends negligibly on the presence or absence of any individual sample in the dataset.
Differential privacy is robust to many forms of attack, such as when the adversary can access side information that nullifies privacy guarantees based on anonymization or mutual information measures (\eg when the adversary can control some of the data vectors $\vx_i$, or can access additional databases that are correlated with the primary database).

For 
\modifiedtext compressive \unmodifiedtext
learning methods, enforcing differential privacy guarantees is as simple as adding well-calibrated noise $\Vec{v}$ to the usual sketch $\empts$, \ie constructing
\begin{equation}
\label{eq:DPsketch}
\psk(\cX) := \empts(\cX) + \Vec{v} ,
\end{equation}
where we find it helpful to explicitly denote the dependence of the dataset $\ds$.
We will assume that the realization $\fmap(\cdot)$ of the random feature map is fixed and publicly known, in contrast to other approaches like \cite{testa2019compressed,rane2013privacy} that use linear mixing matrices as encryption keys to ensure privacy preservation.
As a result, when we treat $\psk(\cX)$ as random, this is due to the randomness in $\Vec{v}$, not the randomness in $\fmap(\cdot)$ or $\cX$.

\begin{figure}
        \centering
        \includegraphics[width=0.95\textwidth]{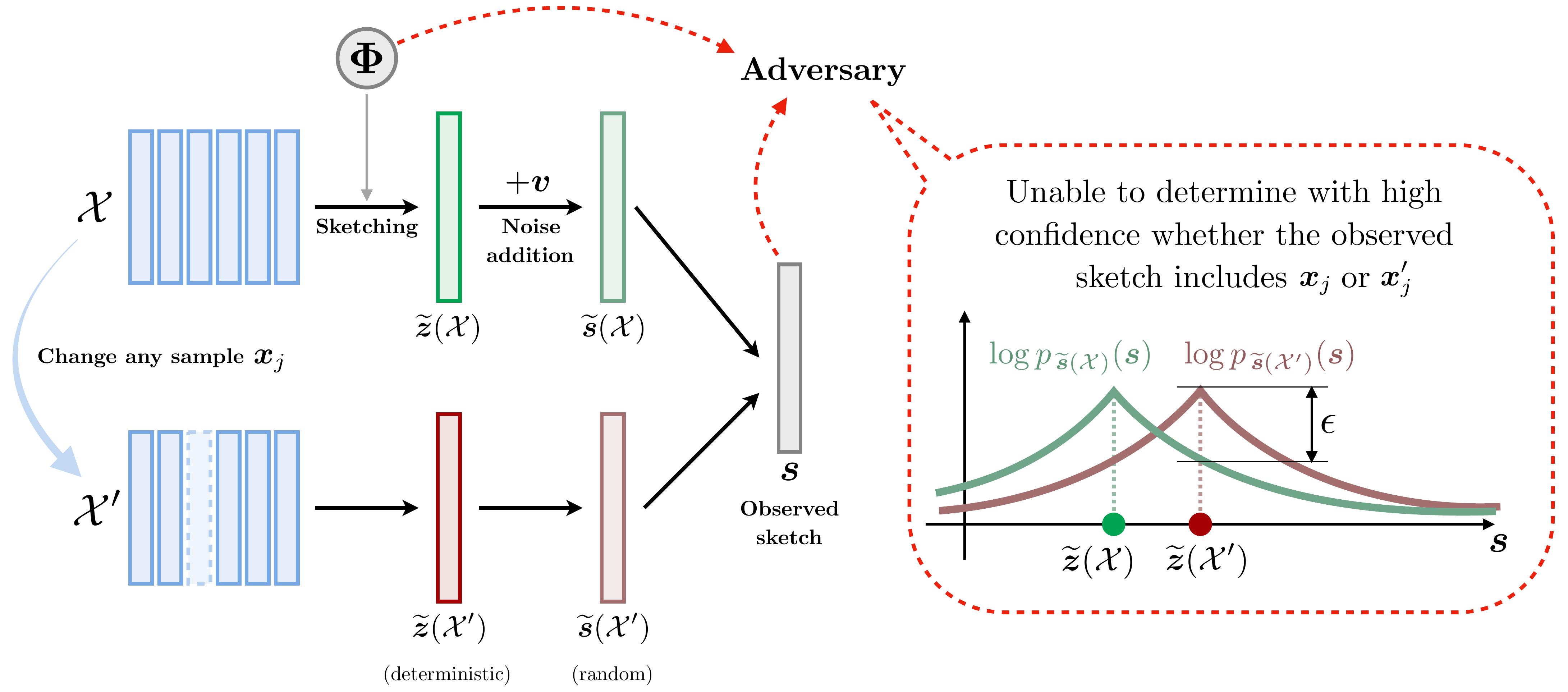}
        \captionof{figure}{When sketching with differential privacy, the output log-density of the sketch $\Vec{s}$ remains close when changing one sample in the dataset (since~\eqref{eq:DPguarantee} is equivalent to $|\log(p_{\psk(\cX)}(\Vec{s})) - \log(p_{\psk(\cX')}(\Vec{s})) | \leq \epsilon$ for all possible $\Vec{s}$). 
        An adversary with knowledge of $\fmap(\cdot)$ and $\Vec{s}$---as symbolized by the red arrows---could then hardly decide whether a given sample $\Vec{x}_j$ was used to compute the sketch or not.
        }
        \label{fig:DP}
\end{figure}

Formally, the sketching mechanism $\psk(\cdot)$ is said to be \emph{$\epsilon$-differentially private} \cite{dwork2014b} if, for any dataset $\cX = \{\vx_i\}_{i=1}^n$ and ``neighboring'' dataset $\cX' = \cX \setminus \{\vx_j\}$ $ \cup \{\vx_j'\}$ that replaces the individual sample $\vx_j$ by another sample $\vx_j'$, and for any possible sketch outcome $\Vec{s}$, we have that
\begin{equation}
\label{eq:DPguarantee}
 \exp(-\epsilon) \leq \frac{p_{\psk(\cX)}(\Vec{s})}{p_{\psk(\cX')}(\Vec{s})} \leq \exp(\epsilon).
\end{equation}
Here, $\epsilon > 0$ plays the role of a privacy level: smaller $\epsilon$ implies a stronger privacy guarantee.
In words, \eqref{eq:DPguarantee} says that, when $\epsilon$ is small, the densities of $\psk(\cX)$ and $\psk(\cX')$ are almost indistinguishable, as depicted in \Cref{fig:DP}. 

The condition~\eqref{eq:DPguarantee} can be interpreted as bounding a ``likelihood ratio'' \cite{poor2013introduction}, a familiar quantity in signal processing.
Consider two hypotheses: one that the dataset equals $\ds$ (\ie includes $\vx_j$), and the other that the dataset equals $\ds'$ (\ie includes $\vx_j'$ instead).
Then $p_{\psk(\cX)}(\Vec{s})$ would be the likelihood of observing private sketch $\Vec{s}$ under the first hypothesis, while $p_{\psk(\cX')}(\Vec{s})$ would be the same for the second hypothesis.
Say an adversary wanted to detect whether or not the dataset contains $\vx_j$. 
By appropriately thresholding the likelihood ratio $p_{\psk(\cX)}(\Vec{s})/p_{\psk(\cX')}(\Vec{s})$, one can obtain hypothesis tests that are optimal from various perspectives (\eg Bayes, minimax, Neyman-Pearson) \cite[Ch.~2]{poor2013introduction}.  Thus,
when \eqref{eq:DPguarantee} holds with small $\epsilon$, it is fundamentally difficult for an adversary to determine whether $\vx_j$ or $\vx_j'$ was present in the sketch.
Even if the adversary had non-trivial prior knowledge of the true hypothesis (as in so-called ``linkage attacks,'' which make use of a second public dataset to which the target user contributed), \eqref{eq:DPguarantee} implies that---for any method---the probability of recovering the true hypothesis from the sketch is only slightly higher than that which is achievable \emph{without} observing the sketch.

To ensure that the noisy sketch \eqref{eq:DPsketch} is differentially private, it is sufficient to draw the noise $\Vec{v}$ as \iid Laplacian with appropriate variance.
The variance needed to achieve a given privacy level $\epsilon$ can be determined by analyzing the so-called \emph{sensitivity} of the noiseless 
sketch, \ie the biggest possible change that can result from removing one sample.
When using the random Fourier feature map~\eqref{eq:RFF}, which generates complex-valued $\empts(\cX)$, it has been established \cite{schellekens2019,chatalic2020CompressiveLearning} that it is sufficient for the real and imaginary components of $\Vec{v}$ to be \iid Laplacian with standard deviation $\sigma_v \propto \frac{m}{n\epsilon}$ .

A weaker form of privacy, known as \emph{approximate differential privacy} or \emph{$(\epsilon,\delta)$-differential privacy} \cite{dwork2014b}, can be attained by adding Gaussian noise $\Vec{v}$ with smaller variance.
For example, with RF features, it is sufficient for the real and imaginary components of $\Vec{v}$ to be \iid Gaussian with standard deviation $\sigma_{v} \propto \frac{\sqrt{m}}{n\epsilon}$. 

The privacy-utility tradeoff facilitates the comparison of different privacy-preserving learning strategies.
For example, given two strategies, one could match the privacy levels $\epsilon$ and compare utilities, or one could match utilities and compare privacy levels.
For 
\modifiedtext compressive \unmodifiedtext 
learning, the utility of interest is the risk (recall~\eqref{eq:learnideal}).


\new{In this section, we focused on differential privacy. 
Other definitions of privacy exist in the literature, such as information-theoretic ones [67]. 
Likewise, cryptographic methods, such as fully homomorphic encryption [68], can be used to transmit and manipulate data in a secure and private manner, although this notion of ``privacy'' is quite different from the former ones. 
Additional work is needed to understand whether compressive learning is ``private'' according to definitions other than differential privacy.}


\section{Perspectives and open challenges}

By averaging well-chosen randomized feature transformations over large training collections, sketching significantly compresses data in a way that facilitates provably accurate yet scalable learning from huge and/or streaming datasets, while simultaneously preserving privacy. 

In this article, we described several approaches to accelerate the sketching process, including feature quantization and the use of randomized fast transforms.
Another approach is to randomly mask each feature vector $\fmap(\vx_i)$ prior to averaging, \ie set a random subset of its components to zero. 
It has been established~\cite{chatalic2020CompressiveLearning} that such random masking does not increase nor decrease the differential privacy level~$\epsilon$.
But it does reduce the need to compute all entries of each feature vector, and thus reduces sketching complexity.
Another promising approach consists of mixed analog-digital sketches, where, \eg optical processing units are used to significantly improve the energy efficiency of the linear stage~\cite{saade2016}. 

When discussing methods \emph{to learn} from a sketch, we focused on optimization-based approaches. 
Although heuristics based on orthogonal matching pursuit~\cite{keriven2017b} \ocite{Jain2011} and approximate message passing~\cite{byrne2019sketched} have been proposed that yield promising empirical results, performance guarantees for these approaches have yet to be established. 
Alternatives, such as total-variation minimization over the space of signed measures~\cite{Candes2014,Denoyelle2018,Boyd2015}, principled greedy methods~\cite{Traonmilin2018}\ocite{Elvira2018}, and gradient flows on systems of particles~\cite{Chizat2018nnflow} could be leveraged to make progress on this front, and black-box optimization could be used to learn from sketches computed by optical processing units.

As for \emph{applications} of 
\modifiedtext compressive \unmodifiedtext 
learning, most of the current literature, and hence most of our article, has focused on unsupervised learning tasks.
Further work is needed to develop 
\modifiedtext compressive \unmodifiedtext 
learning methods for supervised tasks like regression and classification~\cite{ShalevShwartz:2009jm}.
For example, one approach to 
\modifiedtext compressive \unmodifiedtext 
classification was proposed in~\cite{schellekens2018compressive}: for each class $\ell=1,\dots,k$, one computes a sketch $\empts_\ell$ using only the training examples with label $\ell$ (\ie we sketch the $k$ \emph{conditional distributions} of the data). 
From those sketches, one could estimate the conditional densities of each class (using a mixture model, for example), from which a maximum likelihood (ML) or maximum a posteriori (MAP) classifier could be derived. 
Another approach is to perform classification directly in the compressed domain: to an unseen example $\vx'$, we would assign the class $\ell$ that maximizes the correlation $\langle \empts_\ell , \fmap(\vx') \rangle$. 
This strategy can be interpreted ~\cite{schellekens2018compressive} as compressively evaluating a Parzen-window classifier~\cite{pan2008parzen}.
Further work is also needed on unsupervised matrix-factorization tasks like dictionary learning, low-rank matrix completion, and non-negative matrix factorization (NMF)~\cite{udell2016generalized}.

\section*{Acknowledgments}

The authors are grateful to Luc Giffon and the anonymous IEEE-SPM reviewers for their comments on the manuscript.
The authors are funded in part by the Fonds de la Recherche Scientifique (FNRS), the FNRD PDR grant T.0136.20 (Learn2Sense), the Agence Nationale de la Recherche (ANR) under grant ANR-19-CHIA-0009 (AllegroAssai), and the National Science Foundation (NSF) under grant CCF-1955587.



\printbibliography


\end{document}